\documentclass[11pt]{article}

\PassOptionsToPackage{numbers,compress}{natbib}

\usepackage[margin=1in,top=1in]{geometry}
\usepackage{graphicx}
\usepackage{fancyhdr}
\usepackage[dvipsnames]{xcolor}
\usepackage{titlesec}
\usepackage{tcolorbox}
\usepackage{titling}
\tcbuselibrary{skins}
\usepackage{svg}
\usepackage{sfmath}
\usepackage{tikz}
\usetikzlibrary{calc}

\usepackage{natbib}
\usepackage{url}
\usepackage{booktabs}
\usepackage{amsfonts}
\usepackage{amsmath}
\usepackage{amssymb}
\usepackage{mathtools}
\usepackage{amsthm}
\usepackage{newtxtext}
\usepackage{newtxmath}
\usepackage{float}
\usepackage{nicefrac}
\usepackage{microtype}
\usepackage{soul}
\usepackage{enumitem}
\setlist[itemize]{leftmargin=*}
\usepackage{multirow}
\usepackage{caption}
\usepackage{subcaption}
\usepackage[toc,page,header]{appendix}
\usepackage{minitoc}

\usepackage[colorlinks=true,linkcolor=NavyBlue,citecolor=NavyBlue,urlcolor=magenta]{hyperref}
\usepackage[capitalize,noabbrev,nameinlink]{cleveref}

\newcommand{\titlefont}{\sffamily\bfseries}
\newcommand{\emphfont}{\sffamily}
\newcommand{\footerfont}{\sffamily}

\definecolor{qc_blue}{HTML}{2a2aea}
\definecolor{qc_darkblue}{HTML}{020B3F}

\titleformat{\section}
  {\titlefont\Large\bfseries\color{qc_darkblue}}
  {\thesection}{1em}{}
\titlespacing*{\section}{0em}{1em}{.6em}

\newtcolorbox{titlebox}{
  enhanced,
  colback=white,
  boxrule=0pt,
  opacityback=0,
  opacityframe=0,
  width=0.95\textwidth,
  center
}

\makeatletter
\newcommand{\affiliationinfo}[1]{\def\@affiliationinfo{#1}}
\newcommand{\contactinfo}[1]{\def\@contactinfo{#1}}
\newcommand{\authornote}[1]{\def\@authornote{#1}}
\makeatother
\affiliationinfo{}
\contactinfo{}
\authornote{}

\fancypagestyle{titlepage}{
  \fancyhf{}
  \fancyfoot[L]{\footerfont\footnotesize Qualcomm AI Research is an initiative of Qualcomm Technologies, Inc.}
  \fancyfoot[R]{\footerfont\footnotesize\thepage}
  
}

\renewenvironment{abstract}{%
  \setlength{\parskip}{.8em}%
  \par
}{\par}

\newcommand{\ourmethod}{\textsc{ValueDiff}}
\newcommand{\ourmethoddir}{\ourmethod{}-Dir}
\newcommand{\ourmethodnorm}{\ourmethod{}-Norm}

\theoremstyle{plain}
\newtheorem{theorem}{Theorem}[section]
\newtheorem{proposition}[theorem]{Proposition}

\theoremstyle{definition}

\theoremstyle{remark}

\title{ValueDiff: Value-Geometric KV Cache Eviction for Sink-Suppressed LLMs}
\date{}
\author{Junyoung Park\textsuperscript{1}, Jungwook Choi\textsuperscript{2,\textdagger}, Mingu Lee\textsuperscript{1}}
\affiliationinfo{\textsuperscript{1}Qualcomm AI Research \qquad \textsuperscript{2}Hanyang University}
\contactinfo{\{junpark, mingul\}@qualcomm.com, choij@hanyang.ac.kr}
\authornote{\textsuperscript{\textdagger}Contributed to this work while a visiting scholar at Qualcomm AI Research.}

\begin{document}
\thispagestyle{titlepage}

\begin{tikzpicture}[remember picture,overlay]
  \node[anchor=north east,inner sep=0pt]
    at ([xshift=5.8cm,yshift=5.5cm]current page.north east)
    {\includegraphics[width=9cm,angle=-75,origin=c]{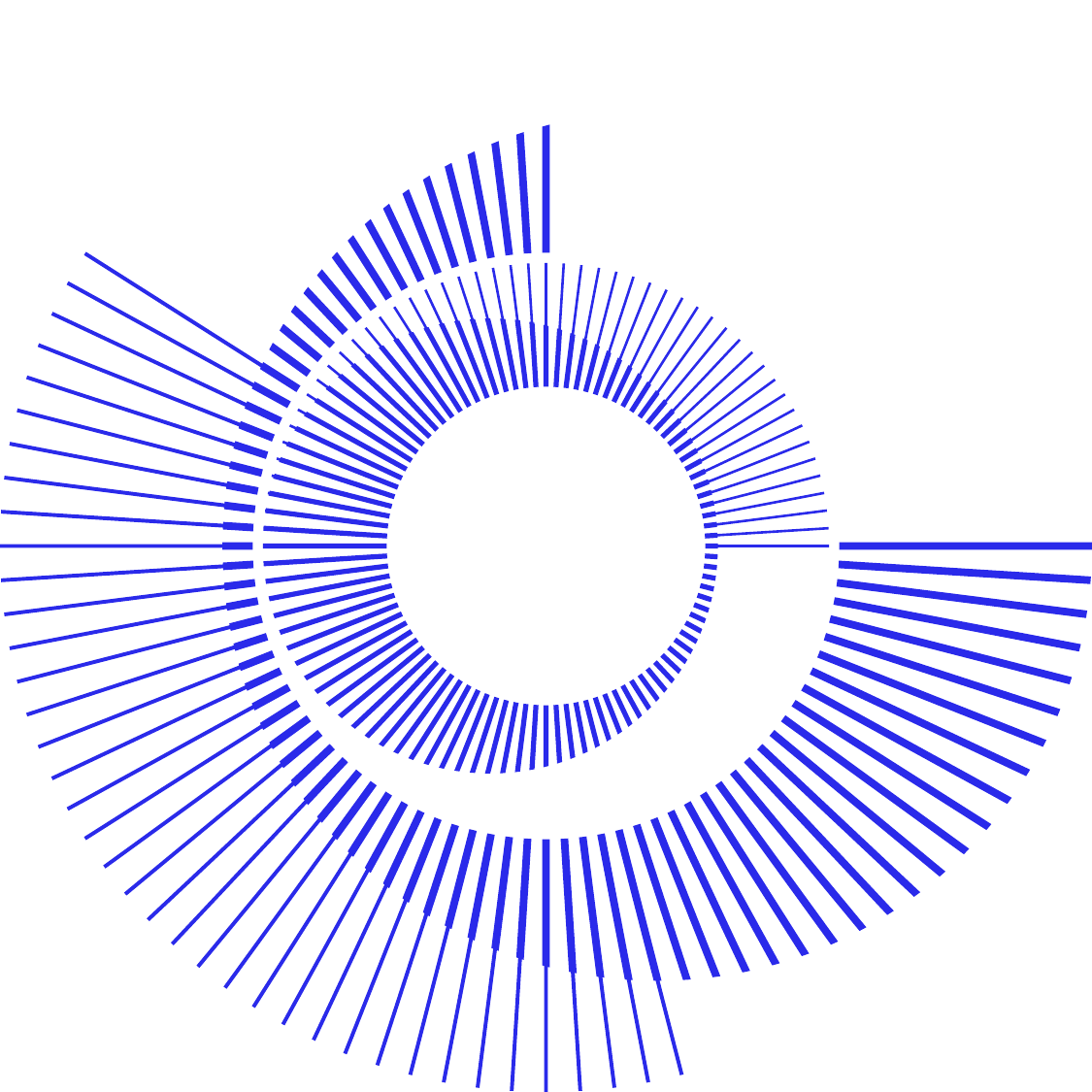}};
\end{tikzpicture}

\begin{figure}[t]
  \vspace*{-1cm}
  \hspace*{-0.6cm}
  {\titlefont\Large\color{qc_blue}Qualcomm AI Research}
  \vspace*{-0.5cm}
\end{figure}

\begin{titlebox}
  {\raggedright\titlefont\Huge\bfseries\color{qc_darkblue}\thetitle\par}
  \vspace{1em}

  \makeatletter
  {\titlefont\color{qc_blue}\@author\par
   \ifx\@affiliationinfo\@empty
   \else
     \vspace{.15em}
     {\emphfont\@affiliationinfo\par}%
   \fi
   \ifx\@contactinfo\@empty
   \else
     \vspace{.05em}
     {\bfseries\emphfont\@contactinfo\par}%
   \fi
   \ifx\@authornote\@empty
   \else
     \vspace{.35em}
     {\footerfont\footnotesize\color{qc_darkblue}\@authornote\par}%
   \fi}
  \makeatother

  \vspace{.5em}
  \begin{abstract}

Modern LLMs with QK-normalization, gated attention, learned attention
sinks, or logit softcapping exhibit weaker persistent attention sinks,
on which existing KV cache eviction methods primarily rely.
We observe that across these models, weaker sinks co-occur with greater
value-vector dispersion relative to key-vector dispersion. Motivated by
this value-side dispersion, we present \ourmethod{}, a value-geometric
eviction that ranks tokens by the L2 deviation of their value vectors
from the cache mean. The same score arises as the minimal-disturbance
eviction under a max-entropy assumption about future attention. We evaluate under fixed cache budgets, with
eviction at every block boundary during prefill and at every decoding
step during generation. On RULER at a tight 2k token budget,
\ourmethod{} retains 88--99\% of dense across seven sink-suppressed
models (best on 6 out of 7). On LongBench at the 4k budget,
\ourmethod{} averages 92\% retention across sink-suppressed models
versus 83\% for the strongest prior baseline. On MATH-500,
\ourmethod{} is the strongest non-dense method on every sink-suppressed
model tested at the 25\% cache budget, outperforming prior
methods by up to $\sim$20 points on gated-attention models. Across all
three benchmarks, value geometry emerges as the more reliable
query-invariant eviction signal for sink-suppressed models.
  \end{abstract}
\end{titlebox}

\section{Introduction}
\label{sec:intro}

Deploying large language models under strict memory constraints (on edge
devices with fixed RAM budgets~\cite{apple_afm,mllmnpu,chenspecnpu}, or in serving systems where
peak GPU memory bounds throughput) requires the KV cache to stay within a
bounded footprint throughout inference, not only during generation.
\emph{KV cache eviction} addresses this by assigning each cached token a
score and discarding the lowest-scoring tokens whenever the cache exceeds a
budget $N$. With long prompts of length $T$, eviction must run during prefill as well. The
prompt is processed in non-overlapping blocks of size $B \ll T$ and the cache
is pruned at each block boundary, a procedure known as
\phantomsection\label{sec:bpp}\emph{block prompt
processing}~\cite{apple_afm,chenspecnpu,keydiff}.
Every eviction decision is \emph{permanent}: a discarded token cannot be
recovered from later context and must remain valid against all future queries.

\paragraph{Attention sinks and architectural trend.}
Attention's spikiness is well-studied as the \emph{attention sink}
phenomenon~\cite{streamingllm,attn_sink_survey,barbero_firsttoken}, in which a small set of
tokens absorbs disproportionate attention mass.
The mechanism is fundamentally key-side: sink tokens receive disproportionate attention
because their key vectors are positively aligned with most query vectors,
yielding consistently high dot products across the majority of query
positions~\cite{keydiff,manifoldkv}.
Prior eviction methods rely on this key-side structure, whether through
attention scores (which measure current key--query alignment) or
key-geometry signals (which use key-vector structure as a proxy for
token importance)~\cite{keydiff,tova,snapkv,h2o,devoto_l2norm,manifoldkv}.
However, recent LLMs incorporating QK-normalization~\cite{qknorm}, gated
attention hybrids~\cite{gatedattn}, learned attention sinks~\cite{gptoss},
and logit softcapping~\cite{gemma2} exhibit lower sink rates: the sink
key vectors no longer maintain persistent alignment with query
vectors nor geometric distinctiveness from content keys
(\cref{tab:sink_dual_mechanism}).
This motivates inspecting eviction signals that do not rely
on key-side information.

\paragraph{Increased value-vector diversity.}
As the key-side signal weakens, value vectors, which form the
``basis'' of the attention output, become the natural place to look
for eviction signals.
Since eviction asks which tokens are redundant and which are distinctive,
we characterize both sides by the spread of value and key vectors
around their respective means, $\sigma_V$ and $\sigma_K$.
Across models spanning four recent architectural families and standard
controls, we find a consistent pattern: models where attention
sinks are weaker tend toward higher $\sigma_V/\sigma_K$, ranging from
key-dominant ($\sigma_V < \sigma_K$) to value-dominant
($\sigma_V > \sigma_K$), as shown in \cref{fig:sink_vs_ratio_polished}.

\paragraph{Value-diversity-based KV cache eviction.}
These observations suggest that value geometry carries more diversity than
key geometry in models with weaker sinks.
We therefore propose \ourmethod{}, which scores each token by
$s_j = \lVert \mathbf{v}_j - \bar{\mathbf{v}} \rVert$ where
$\bar{\mathbf{v}}$ is the cache mean (\cref{eq:vd}).
Low-$s_j$ tokens are near-redundant and evicted first, while high-$s_j$
tokens carry distinctive content and are retained. The same score arises from
symmetry: when future attention is unknown, \ourmethod{} is the eviction that
least disturbs the attention output under a uniform default
(\cref{sec:theory}).

\paragraph{Contributions.}
\begin{itemize}[leftmargin=*,itemsep=2pt,topsep=4pt]
  \item \textbf{Value-geometric eviction.} Across models
    spanning four recent architectural families and standard controls,
    weakened sink behavior co-occurs with value-dominant geometry. This motivates \ourmethod{},
    $s_j = \lVert \mathbf{v}_j - \bar{\mathbf{v}} \rVert$, and a
    pair of endpoint ablations, \ourmethoddir{} and \ourmethodnorm{}, used
    to test whether value information is carried more by direction or by norm
    (\cref{sec:observation,sec:score}).

  \item \textbf{Max-entropy interpretation.} When future attention is unknown,
    \ourmethod{} is the eviction that minimally disturbs the attention output
    under the least-informative assumption, providing a principled derivation of
    the score from a symmetry argument
    (\cref{prop:vdiff_optimum}, \cref{app:attention_flatness}).

  \item \textbf{Empirical validation.} \ourmethod{}
    achieves best or tied-best RULER retention on 6 of the 7 sink-suppressed
    models at the tight 2k-token budget. On LongBench,
    \ourmethod{} averages 92\% retention across the sink-suppressed models at the tighter 4k budget, versus 83\% for the strongest prior baseline
    (\cref{sec:experiments}).
    Further analysis shows that the observed benchmark patterns are
    well-accounted for by a two-axis geometric picture combining $\sigma_V/\sigma_K$
    and key-space alignment (\cref{sec:discussion}).
\end{itemize}

\section{Background}
\label{sec:background}

\subsection{KV Caching and Eviction}

The attention operator~\cite{vaswani_attention} projects input tokens into queries, keys, and values ($\mathbf{Q} = \mathbf{X}\mathbf{W}_Q$, $\mathbf{K} = \mathbf{X}\mathbf{W}_K$, $\mathbf{V} = \mathbf{X}\mathbf{W}_V$) and computes
\begin{equation}\label{eq:attention}
  \mathbf{O} = \text{Softmax}\!\left(\mathbf{Q}\mathbf{K}^\top / \sqrt{d} + \mathbf{M}\right)\mathbf{V} = \mathbf{A}\mathbf{V},
\end{equation}
where $\mathbf{M}$ is the causal mask. During autoregressive generation, previously computed key-value pairs are stored in a \emph{KV cache} $\mathcal{C} = (\mathbf{K}, \mathbf{V})$ and reused. The cache grows linearly with the number of processed tokens and dominates memory for long sequences. Architectural sharing such as grouped-query attention~\cite{gqa} reduces per-token cache size but leaves the linear-in-length growth unchanged, and IO-efficient attention kernels~\cite{flashattention} that can accelerate the computation do not reduce the cache footprint.

To bound memory usage, an eviction policy $\pi_N(\mathcal{C})$ removes tokens whenever the cache size exceeds a budget $N$. Given a scoring function $s(\cdot)$, eviction retains the $N$ highest-scoring tokens:
\begin{equation}\label{eq:eviction}
  \mathcal{S} = \text{top-}N(s(\mathcal{C})), \quad
  \mathbf{K}' = \text{gather}(\mathbf{K}, \mathcal{S}), \quad
  \mathbf{V}' = \text{gather}(\mathbf{V}, \mathcal{S}).
\end{equation}
Attention-score methods (H\textsubscript{2}O~\cite{h2o}, TOVA~\cite{tova},
SnapKV~\cite{snapkv}, KVZip~\cite{kvzip}) define $s(\cdot)$ using attention
weights, either from observed queries or from reconstruction prompts.
Key-geometric methods such as \textsc{KeyDiff}~\cite{keydiff} and
ManifoldKV~\cite{manifoldkv} use query-invariant properties of the keys.

\subsection{Block Prompt Processing}
\label{sec:block_prompt}

Standard KV cache eviction assumes the full prompt is processed before eviction begins. This requires materializing the entire KV cache during prefill, which may exceed the memory budget on resource constrained devices. \emph{Block prompt processing}~\cite{keydiff,caote} addresses this by segmenting the input $\mathbf{X} = [\mathbf{X}_0, \mathbf{X}_1, \ldots, \mathbf{X}_{m-1}]$ into non-overlapping blocks of size $B$ and applying eviction after each block.
\begin{equation}\label{eq:block_processing}
  \mathcal{C}_i \leftarrow (\,[\mathbf{K}_{i-1} \| \mathbf{k}_{Bi:B(i+1)-1}],\; [\mathbf{V}_{i-1} \| \mathbf{v}_{Bi:B(i+1)-1}]\,), \quad
  \mathcal{C}_i \leftarrow \pi_N(\mathcal{C}_i)
\end{equation}
with $\mathcal{C}_{-1} = \emptyset$, for $i = 0, \ldots, m-1$. By applying $\pi_N$ at every block boundary, this bounds the cache to $N$ tokens during prefill as well as generation. We note that block prompt processing with $B = T$ (the full prompt length) recovers standard eviction, and $B = 1$ corresponds to token-by-token generation.

\begin{figure}[t]
\centering
\includegraphics[width=0.82\linewidth]{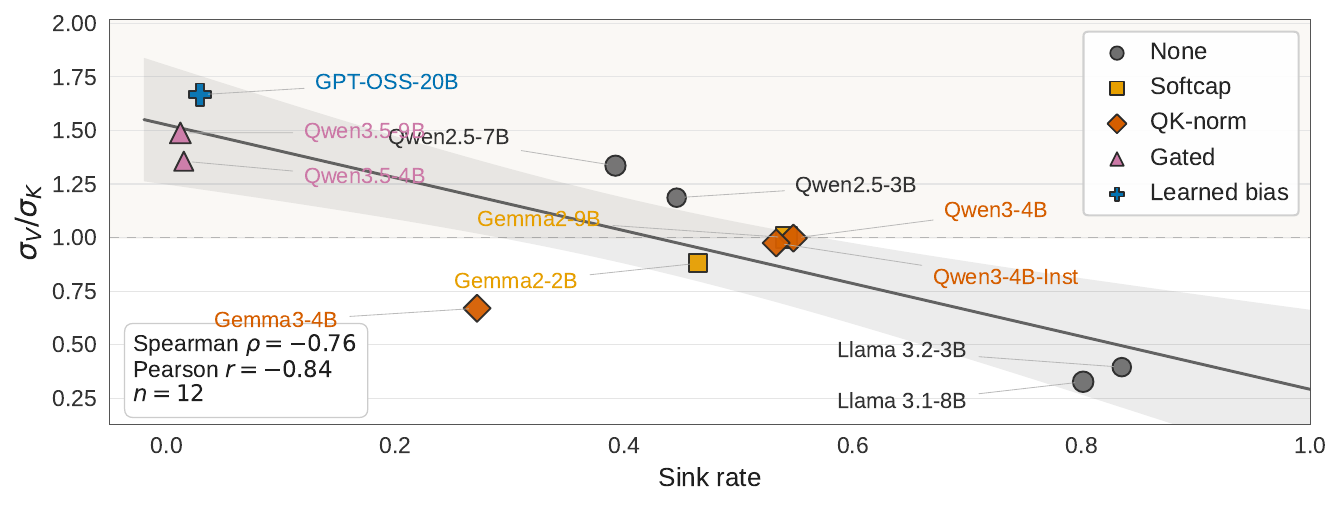}
\caption{Sink rate versus $\sigma_V/\sigma_K$. Color and marker encode
  suppression category; shaded band = 95\% conditional-mean CI; region above
  $\sigma_V/\sigma_K=1$ marks value-dominant geometry}
\label{fig:sink_vs_ratio_polished}
\vspace{-6pt}
\end{figure}

\section{\ourmethod{} for Value-Geometric KV Cache Eviction}
\label{sec:valuediff}

\subsection{Sink Rate and Value Geometry}
\label{sec:observation}\label{sec:confirmation}

We measure the maximum sink rate and the per-token L2 dispersion of key and
value vectors across models spanning four recent architectural families and
standard controls
(\cref{fig:sink_vs_ratio_polished}; full statistics in \cref{tab:kv_diversity_full}).
As a general trend, models with lower sink rates tend to exhibit greater value-vector dispersion relative to keys.
For head $h$ at layer $\ell$ with $n$ tokens, let
$\sigma_K^{(\ell,h)} = \mathrm{std}_i(\lVert k_i^{(\ell,h)} - \bar{k}^{(\ell,h)} \rVert_2)$
and $\sigma_V^{(\ell,h)} = \mathrm{std}_i(\lVert v_i^{(\ell,h)} - \bar{v}^{(\ell,h)} \rVert_2)$,
where $\bar{k}^{(\ell,h)} = \tfrac{1}{n}\sum_i k_i^{(\ell,h)}$.
Each model's position in \cref{fig:sink_vs_ratio_polished} uses the mean of
$\sigma_K^{(\ell,h)}$ and $\sigma_V^{(\ell,h)}$ across all layers and heads.
Sink rate---the fraction of heads where the leading token exceeds weight
$\tau{=}0.3$~\cite{ruscio2025sinking}---ranges from 0.86--0.90 on Llama~3
to near zero on Qwen3.5 and GPT-OSS-20B. Models with lower sink rates
tend to show higher $\sigma_V/\sigma_K$ (\cref{fig:sink_vs_ratio_polished}).
Qwen2.5 is a notable exception with high
per-layer geometric heterogeneity; we discuss this separately in
\cref{app:per_layer_geometry}.

The sink-rate observation directly challenges sink-preserving eviction. When
leading-token sink rates approach zero, fixed-sink policies such as
StreamingLLM~\cite{streamingllm} have less reason to privilege the earliest
cache positions. For attention-score methods such as TOVA~\cite{tova} and
SnapKV~\cite{snapkv}, the implication is subtler. Persistent high-mass sink positions provide a stable shortcut; as those
sinks weaken, attention-score methods fall back on query-specific rankings
under block prompt processing, where evictions are permanent and future
queries are unavailable. Attention-ranking overlap is consistently modest
(top-256 Jaccard $\approx 0.4$; \cref{app:attn_rank_stability}), making
this fallback an unreliable proxy for future cache utility.
The value-geometry observation challenges key-only geometric eviction.
Since eviction asks which tokens are redundant and which are distinctive,
the rising $\sigma_V/\sigma_K$ points to value geometry as the stronger
query-invariant signal for this discrimination. We revisit this picture after the empirical results
through an additional key-space alignment measure that further characterizes the geometric conditions (\cref{sec:discussion}).

\subsection{\ourmethod{} Score}
\label{sec:score}

We define the \ourmethod{}
score as
\begin{equation}\label{eq:vd}
  s_j \;=\; \lVert \mathbf{v}_j - \bar{\mathbf{v}} \rVert, \qquad
  \bar{\mathbf{v}} \;=\; \tfrac{1}{n}\textstyle\sum_i \mathbf{v}_i,
\end{equation}
where $n$ is the number of cached tokens.
Intuitively, a high $s_j$ marks a token as geometrically distinctive and
worth retaining; a low $s_j$ marks it as near-redundant and safely discarded.

\paragraph{Two axes of content.}
Expanding \cref{eq:vd} via the law of cosines isolates two distinct
geometric quantities,
\begin{equation}\label{eq:decomp}
  s_j^2 \;=\; \lVert \mathbf{v}_j \rVert^2
       - 2\lVert \mathbf{v}_j \rVert\lVert \bar{\mathbf{v}} \rVert \cos\theta_j
       + \lVert \bar{\mathbf{v}} \rVert^2,
\end{equation}
where $\theta_j$ is the angle between $\mathbf{v}_j$ and $\bar{\mathbf{v}}$.
This L2 score is shaped by \textbf{magnitude} ($\lVert \mathbf{v}_j \rVert$,
tokens with large value norms) and \textbf{direction} ($1-\cos\theta_j$,
tokens pointing away from the mean regardless of norm).
To understand the role of each axis, we further evaluate two \ourmethod{}
variants. \ourmethoddir{} keeps only the directional score
$1-\cos(\mathbf{v}_j,\bar{\mathbf{v}})$, and \ourmethodnorm{} keeps
only the magnitude score $\lVert\mathbf{v}_j\rVert$. These variants ask
whether value information is carried more by direction or by norm.

\subsection{A Max-Entropy Interpretation of \ourmethod{}}
\label{sec:theory}

Eviction ultimately aims to preserve information that may be useful under
unknown future attention distributions. Modeling those distributions is
challenging in both block prompt processing and standard full-prefill eviction.
We therefore ask which token is safest to remove under the least-informative
assumption about future attention.

\paragraph{Expected output disturbance.} The attention output under a future attention vector
$\boldsymbol{\alpha} \in \Delta^{n-1}$ is
$\mathbf{o}(\boldsymbol{\alpha}) = \sum_i \alpha_i \mathbf{v}_i$. A natural eviction criterion is the expected disturbance to the attention output
when token $j$ is removed and the remaining attention weights are renormalized
over the survivors.
\begin{equation}\label{eq:recon_obj}
  j^\star \;=\; \arg\min_j\;
  \mathbb{E}_{\boldsymbol{\alpha}\sim P}\!\left[
    \bigl\lVert \mathbf{o}(\boldsymbol{\alpha}) - \mathbf{o}_{\setminus j}(\boldsymbol{\alpha}) \bigr\rVert^2
  \right],
  \qquad
  \mathbf{o}_{\setminus j}(\boldsymbol{\alpha})
  \;=\; \sum_{i\neq j} \tfrac{\alpha_i}{1-\alpha_j}\, \mathbf{v}_i,
\end{equation}
where $P$ is a distribution on $\Delta^{n-1}$ encoding our belief about the
future attention vector. With no information distinguishing tokens at eviction
time, the symmetric choice is the uniform attention vector
$\boldsymbol{\alpha}^\star = (1/n, \ldots, 1/n)$, the unique point on
$\Delta^{n-1}$ invariant under permutations of token indices.

\paragraph{Reduction at $\boldsymbol{\alpha}^\star$.}
Substituting $\boldsymbol{\alpha} = \boldsymbol{\alpha}^\star$,
the attention output collapses to the per-head cache
mean, and uniform redistribution over the remaining $n-1$ tokens gives the
post-eviction output.
\begin{equation}\label{eq:uniform_outputs}
  \mathbf{o}^\star \;=\; \bar{\mathbf{v}},
  \qquad
  \mathbf{o}^\star_{\setminus j}
  \;=\; \sum_{i\neq j} \tfrac{1}{n-1}\, \mathbf{v}_i
  \;=\; \frac{n\, \bar{\mathbf{v}} - \mathbf{v}_j}{\,n-1\,}.
\end{equation}
Evaluating the integrand of \cref{eq:recon_obj} at $\boldsymbol{\alpha} = \boldsymbol{\alpha}^\star$ reduces the objective to
\begin{equation}\label{eq:recon_error}
  \bigl\lVert \mathbf{o}^\star - \mathbf{o}^\star_{\setminus j} \bigr\rVert^2
  \;=\; \frac{1}{(n-1)^2}\, \bigl\lVert \mathbf{v}_j - \bar{\mathbf{v}} \bigr\rVert^2,
\end{equation}
whose minimizer over $j$ is exactly the \ourmethod{} score.

\begin{proposition}[\ourmethod{} as the symmetric optimum]\label{prop:vdiff_optimum}
At the symmetric attention vector
$\boldsymbol{\alpha} = \boldsymbol{\alpha}^\star = (1/n,\ldots,1/n)$,
the eviction objective in \cref{eq:recon_obj} is minimized at
\begin{equation*}
  j^\star \;=\; \arg\min_j \bigl\lVert \mathbf{v}_j - \bar{\mathbf{v}} \bigr\rVert,
  \qquad
  \bar{\mathbf{v}} \;=\; \tfrac{1}{n}\textstyle\sum_i \mathbf{v}_i.
\end{equation*}
\end{proposition}

\paragraph{Remark.}
The derivation above uses the symmetric reference point
$\boldsymbol{\alpha}^{\star}$. The same ranking also arises from the
max-entropy version of the reconstruction objective under an unconstrained
Dirichlet$(1,\ldots,1)$ distribution over future attention
(\cref{app:proof_prop_vdiff}). The relevance of this reference point grows
as attention distributions flatten; \cref{app:attention_flatness} formalizes
how the distance from the symmetric limit bounds the approximation error.

\section{Experiments}
\label{sec:experiments}

\begin{table}[t]
\centering
\caption{RULER accuracy retention (\% of dense, $\uparrow$; ctx=8192).
  Each model has two budget rows (2048, 4096). \textbf{Bold} marks the best non-dense strategy per budget,
  and \underline{underline} marks the second best.}
\label{tab:ruler_main}
\resizebox{\linewidth}{!}{%
\begin{tabular}{llcccccccccc}
\toprule
 & & \multicolumn{1}{c}{\textit{Sink}} & \multicolumn{2}{c}{\textit{Attn}} & \multicolumn{3}{c}{\textit{Key Geometry}} & \multicolumn{1}{c}{\textit{Attn$\times$Val.}} & \multicolumn{3}{c}{\textit{Value Geometry}} \\
\cmidrule(lr){3-3}\cmidrule(lr){4-5}\cmidrule(lr){6-8}\cmidrule(lr){9-9}\cmidrule(lr){10-12}
Model & Bud. & Streaming & TOVA & SnapKV & KNorm & KeyDiff & ManifKV & F-CAOTE & V-Norm & V-Dir & \ourmethod{} \\
\midrule
\multicolumn{12}{l}{\textit{Standard}} \\
\multirow{2}{*}{Llama 3.1-8B} & 2k & 40.1 & 62.0 & 75.5 & \textbf{91.8} & \underline{89.6} & 60.9 & 78.8 & 53.1 & 88.7 & 88.1 \\
 & 4k & 60.4 & 85.6 & 85.0 & \textbf{95.5} & \underline{95.2} & 84.9 & 90.0 & 82.9 & 94.1 & 95.0 \\
\multirow{2}{*}{Llama 3.2-3B} & 2k & 40.2 & 59.7 & 72.2 & 79.6 & \textbf{92.9} & 40.5 & 72.3 & 44.5 & \underline{89.3} & 88.8 \\
 & 4k & 60.6 & 82.4 & 82.3 & 94.1 & \textbf{97.4} & 89.9 & 87.9 & 78.7 & \underline{96.5} & 94.7 \\
\multirow{2}{*}{Qwen2.5-3B} & 2k & 18.0 & 22.1 & 23.7 & 31.4 & \underline{39.8} & 34.4 & 28.8 & 19.6 & 39.7 & \textbf{41.1} \\
 & 4k & 22.0 & 48.5 & 39.6 & 60.7 & 63.6 & 62.7 & 55.5 & 43.9 & \underline{63.7} & \textbf{65.4} \\
\multirow{2}{*}{Qwen2.5-7B} & 2k & 10.9 & 14.0 & 18.7 & 23.1 & \underline{31.9} & 13.1 & 20.3 & 19.3 & \textbf{32.4} & 21.8 \\
 & 4k & 14.5 & 36.8 & 43.1 & 52.3 & \underline{56.2} & 46.7 & 42.6 & 36.6 & \textbf{56.6} & 46.7 \\
\midrule
\multicolumn{12}{l}{\textit{Logit softcapping}} \\
\multirow{2}{*}{Gemma2-2B} & 2k & 48.7 & 59.2 & 64.5 & 79.8 & \textbf{96.4} & 50.6 & 71.1 & 54.9 & \underline{93.7} & 89.6 \\
 & 4k & 63.7 & 82.8 & 80.0 & 86.5 & \textbf{98.7} & 57.6 & 86.2 & 61.5 & \underline{98.1} & 95.7 \\
\multirow{2}{*}{Gemma2-9B} & 2k & 52.3 & 66.7 & 81.7 & 59.5 & \underline{99.0} & 66.0 & 72.4 & 54.2 & 97.8 & \textbf{99.4} \\
 & 4k & 66.5 & 69.7 & 85.8 & 67.9 & 99.3 & 80.6 & 77.8 & 72.2 & \underline{99.6} & \textbf{100.0} \\
\midrule
\multicolumn{12}{l}{\textit{QK-normalization}} \\
\multirow{2}{*}{Qwen3-4B} & 2k & 42.4 & 44.5 & 52.8 & 40.5 & 89.6 & \underline{92.7} & 71.7 & 89.7 & 89.8 & \textbf{95.2} \\
 & 4k & 63.0 & 81.0 & 78.6 & 69.8 & 95.1 & \underline{98.0} & 91.6 & \textbf{98.4} & 97.1 & \underline{98.0} \\
\multirow{2}{*}{Gemma3-4B} & 2k & 40.6 & 54.7 & 66.2 & 11.9 & 81.7 & \underline{87.1} & 69.1 & 59.4 & 86.2 & \textbf{87.6} \\
 & 4k & 63.0 & 76.1 & 80.5 & 19.9 & 91.9 & 94.6 & 91.7 & 94.5 & \underline{95.0} & \textbf{96.1} \\
\midrule
\multicolumn{12}{l}{\textit{Gated attention}} \\
\multirow{2}{*}{Qwen3.5-4B} & 2k & 42.0 & 42.0 & 62.6 & 59.7 & 93.3 & 92.5 & 67.6 & 93.7 & \underline{94.7} & \textbf{97.6} \\
 & 4k & 61.4 & 66.5 & 82.2 & 96.0 & 96.6 & 97.0 & 84.1 & 98.7 & \underline{99.5} & \textbf{99.7} \\
\multirow{2}{*}{Qwen3.5-9B} & 2k & 42.9 & 42.0 & 77.2 & 24.2 & 91.9 & 92.1 & 66.4 & 86.9 & \underline{95.4} & \textbf{95.6} \\
 & 4k & 61.6 & 61.6 & 84.3 & 70.2 & 96.2 & 96.7 & 81.8 & \textbf{98.4} & \underline{98.2} & \underline{98.2} \\
\midrule
\multicolumn{12}{l}{\textit{Learned Sink}} \\
\multirow{2}{*}{GPT-OSS-20B} & 2k & 39.3 & 54.2 & 80.2 & 64.1 & 88.9 & 68.5 & 75.7 & 50.2 & \underline{90.1} & \textbf{91.3} \\
 & 4k & 62.8 & 85.0 & 87.1 & 88.0 & 93.1 & \underline{94.9} & 90.1 & 86.6 & 93.2 & \textbf{97.8} \\
\bottomrule
\end{tabular}}
\end{table}

\paragraph{Evaluation protocol.}
We evaluate under \emph{block prompt processing} (\cref{eq:block_processing}) with a chunk size of $B=128$ combined with decode-time token eviction, following the same protocol as~\cite{keydiff,caote,mllmnpu,chenspecnpu}.
This is strictly harder than the full-prefill setting assumed by most token
eviction work because decisions are made with only partial context available,
and discarded tokens cannot
be recovered when later chunks arrive.

\paragraph{Models.}
Models span four architectural mechanisms. \textbf{QK-normalization}
(per-head RMSNorm on queries and keys; Qwen3-4B, Gemma3-4B); \textbf{gated attention} (Qwen3.5-4B, 9B);
\textbf{learned attention sink} (per-head sink logit appended to the attention logits;
GPT-OSS-20B); and \textbf{logit softcapping} (tanh-bounded attention logits;
Gemma2-2B, 9B). Standard architectures (Llama~3.1-8B,
3.2-3B~\cite{llama3}; Qwen2.5-3B, 7B~\cite{qwen25}) serve as controls.

\paragraph{Baselines.}
Baselines span the main eviction signal categories.
\textbf{Attention sink.} StreamingLLM~\cite{streamingllm} (fixed sink + recent window).
\textbf{Attention score.} TOVA~\cite{tova} and SnapKV~\cite{snapkv}.
\textbf{Key geometry.} KeyNorm~\cite{devoto_l2norm} ranks tokens by
key-vector norm, preferring smaller norms; \textsc{KeyDiff}~\cite{keydiff}
scores cosine distance from the mean key; and ManifoldKV~\cite{manifoldkv}
scores Euclidean distance from a key centroid. These key-side baselines
match the radial, directional, and Euclidean value-side signals tested by
\ourmethodnorm{}, \ourmethoddir{}, and \ourmethod{}, respectively, except
that KeyNorm uses the opposite radial preference from \ourmethodnorm{}.
\textbf{Attention $\times$ value geometry.} FastCAOTE~\cite{caote}
($\tfrac{\alpha_j}{1-\alpha_j}\cdot\lVert\mathbf{v}_j-\bar{\mathbf{v}}\rVert$),
which combines value geometry with an attention multiplier; comparing it with \ourmethod{}
directly isolates the effect of that multiplier.
\textbf{Value geometry.} \ourmethoddir{} (direction-only), \ourmethod{} (L2),
and \ourmethodnorm{} (magnitude-only).
KVZip~\cite{kvzip} is excluded: its doubled-length prefill per eviction step
is prohibitive under block prompt processing. All methods keep the first and 128 most recent tokens, which count toward the cache budget.

\paragraph{Benchmarks.}
We cover both long-context prefill and long decode.
\textbf{RULER}~\cite{ruler} evaluates long-context capability in
controlled 8k-context tasks spanning retrieval (NIAH~\cite{niah}), variable
tracking, aggregation, and multi-hop QA, while \textbf{LongBench}~\cite{longbench}
tests transfer to 21 heterogeneous document tasks with contexts reaching up to 65k tokens. \textbf{MATH-500}~\cite{hendrycks_math,lightman2023lets}
stresses decode-time cache growth during chain-of-thought reasoning.

\subsection{RULER}
\label{sec:ruler_results}

On RULER, we evaluate long-context capability at 8k context under tight cache
budgets of 2k and 4k tokens. \cref{tab:ruler_main} reports retention versus
each model's dense score across 13 RULER tasks. The results split across three geometry regimes consistent with
\cref{sec:observation}. On value-dominant models (Qwen3.5, GPT-OSS-20B;
$\sigma_V/\sigma_K > 1.3$), \ourmethod{} is the strongest. On mixed-geometry
models ($\sigma_V/\sigma_K \approx 1$; Qwen3, Gemma2, Gemma3), \ourmethod{}
generally performs best, with key-geometry methods such as KeyDiff and
ManifoldKV remaining competitive. In this regime neither signal has a clear
geometric advantage; \ourmethod{} leads on most models here, with Gemma2-2B
the exception where KeyDiff leads---consistent with its $\sigma_V/\sigma_K{=}0.88$
placing it toward the key-dominant side (\cref{sec:discussion}).
On key-dominant non-suppressed controls ($\sigma_V/\sigma_K < 0.5$; Llama),
KeyDiff and KeyNorm lead. Qwen2.5 is an outlier: both value- and
key-geometry methods degrade substantially, consistent with the high
per-layer geometry variance discussed in \cref{app:per_layer_geometry}.

\paragraph{Sink and attention baselines.}
The non-geometric baselines expose two different failure modes. Streaming's
fixed-window design fails broadly at tight budgets regardless of sink behavior;
on sink-suppressed models it additionally loses the stable BOS anchor that
partially compensates on sink-dominated architectures. TOVA and SnapKV face
a different problem. The attention scores available at a block boundary must
predict token importance for later blocks and decoding, and the low rank overlap
in \cref{app:attn_rank_stability} suggests that this proxy is unreliable.

\paragraph{Key vs.\ value geometry.}
Matched key/value pairs consistently favor the value side on sink-suppressed
models: \ourmethod{} matches or improves on ManifoldKV (centroid), and
\ourmethoddir{} improves over KeyDiff (directional). The norm-only comparison
is less diagnostic because both KeyNorm and \ourmethodnorm{} underperform
their directional or L2 counterparts, indicating norm alone is not dominant.
Overall, value geometry is the more robust query-invariant signal; we examine
the underlying geometric conditions in \cref{sec:discussion}.

\paragraph{Comparison with FastCAOTE.}
FastCAOTE keeps the same centroid-distance signal as \ourmethod{} but weights
it by observed attention. The multiplier consistently hurts relative to
\ourmethod{} (\cref{tab:ruler_main}), matching the broader failure mode of
attention-based eviction: observed attention is query-local, not a stable
estimate of future cache utility (\cref{app:attn_rank_stability}).

\paragraph{Direction and magnitude ablations.}
Within the value-side family, \ourmethoddir{} is often close to the default
score, suggesting that much of the useful signal is directional. However,
\ourmethodnorm{} is much less reliable on its own, and the default L2 score is
the most stable variant overall. This suggests that magnitude is not a strong
standalone eviction criterion, but it helps calibrate the directional signal
when combined through centroid distance.

\begin{table}[t]
\centering\small
\caption{LongBench accuracy retention (\% of dense, $\uparrow$) at \textbf{4k\,/\,8k}
  budgets.
  \textbf{Bold} marks the best non-dense strategy by average retention over
  the two reported budgets for each model.}
\label{tab:longbench}
\resizebox{\linewidth}{!}{%
\begin{tabular}{lcccccccccc}
\toprule
 & \multicolumn{1}{c}{\textit{Sink}} & \multicolumn{2}{c}{\textit{Attn}} & \multicolumn{3}{c}{\textit{Key Geometry}} & \multicolumn{1}{c}{\textit{Attn$\times$Val.}} & \multicolumn{3}{c}{\textit{Value Geometry}} \\
\cmidrule(lr){2-2}\cmidrule(lr){3-4}\cmidrule(lr){5-7}\cmidrule(lr){8-8}\cmidrule(lr){9-11}
Model & Streaming & TOVA & SnapKV & KeyNorm & KeyDiff & ManifoldKV & F-CAOTE & V-Norm & V-Dir & \ourmethod{} \\
\midrule
\multicolumn{11}{l}{\textit{Standard}} \\
\quad Llama3.1-8B & 55.0\,/\,74.4 & 69.0\,/\,88.4 & 70.6\,/\,91.0 & 73.5\,/\,90.3 & \textbf{82.6\,/\,93.6} & 71.7\,/\,91.9 & 71.4\,/\,90.0 & 65.1\,/\,85.2 & 79.7\,/\,91.5 & 73.1\,/\,88.0 \\
\quad Llama3.2-3B & 70.0\,/\,79.4 & 74.8\,/\,85.5 & 75.9\,/\,85.6 & 76.6\,/\,86.3 & \textbf{85.4\,/\,89.2} & 76.9\,/\,86.8 & 76.4\,/\,85.3 & 69.6\,/\,82.2 & 79.7\,/\,87.5 & 75.9\,/\,85.1 \\
\midrule
\multicolumn{11}{l}{\textit{QK-normalization}} \\
\quad Qwen3-4B   & 78.0\,/\,89.4 & 81.3\,/\,92.0 & 83.8\,/\,93.7 & 71.9\,/\,87.4 & 85.0\,/\,94.6 & 89.5\,/\,97.1 & 84.0\,/\,94.9 & 95.2\,/\,98.8 & 91.2\,/\,97.4 & \textbf{95.3\,/\,98.7} \\
\quad Gemma3-4B  & 72.1\,/\,83.7 & 77.0\,/\,90.5 & 72.7\,/\,86.6 & 49.5\,/\,58.2 & 76.7\,/\,90.9 & 73.0\,/\,79.8 & 78.2\,/\,92.5 & 82.6\,/\,93.0 & \textbf{90.5\,/\,96.5} & 88.1\,/\,94.5 \\
\midrule
\multicolumn{11}{l}{\textit{Gated attention}} \\
\quad Qwen3.5-4B & 77.7\,/\,87.9 & 84.0\,/\,91.2 & 83.1\,/\,93.0 & 79.4\,/\,92.0 & 78.6\,/\,90.7 & 80.5\,/\,93.1 & 86.2\,/\,94.1 & \textbf{92.8\,/\,97.6} & 88.4\,/\,95.8 & 92.2\,/\,95.5 \\
\quad Qwen3.5-9B & 74.1\,/\,86.7 & 84.6\,/\,95.2 & 85.6\,/\,95.1 & 82.2\,/\,92.4 & 81.3\,/\,92.9 & 80.9\,/\,92.8 & 87.1\,/\,95.7 & 92.4\,/\,96.7 & 90.0\,/\,95.5 & \textbf{92.1\,/\,97.2} \\
\midrule
\multicolumn{11}{l}{\textit{Learned Sink}} \\
\quad GPT-OSS-20B & 60.7\,/\,79.1 & 75.3\,/\,91.3 & 74.4\,/\,90.2 & 25.3\,/\,28.5 & 83.4\,/\,94.5 & 31.0\,/\,33.2 & 77.8\,/\,93.6 & 81.9\,/\,96.3 & 84.4\,/\,95.9 & \textbf{91.0\,/\,96.6} \\
\bottomrule
\end{tabular}}
\end{table}

\subsection{LongBench}
\label{sec:longbench}

Through RULER, we confirmed that the value-geometry advantage holds in
controlled, synthetic long-context conditions. In LongBench, we further
evaluate the methods on 21 heterogeneous tasks spanning bilingual QA, multi-document summarization, code completion, and
open-ended retrieval. In this more diverse setting, \ourmethod{} retains
88--95\% of dense performance on sink-suppressed models at the tighter 4k budget (see \cref{tab:longbench}). Note that logit softcapping models (Gemma2-2B and Gemma2-9B) are excluded from the evaluation as their trained context length is 8k \cite{gemma2}. Qwen2.5 models are also excluded: severe per-layer geometry heterogeneity causes all eviction strategies to degrade substantially regardless of signal type (see \cref{app:per_layer_geometry}), making their inclusion uninformative for the geometric comparison.

Magnitude-based key selection (KeyNorm, ManifoldKV) shows architecture-dependent failures in LongBench. On Gemma3-4B (QK-norm), KeyNorm collapses to 49.5\% retention at the 4k budget; on GPT-OSS-20B (learned sink), both KeyNorm (25.3\%) and ManifoldKV (31.0\%) fail catastrophically.
KeyDiff, insensitive to key magnitude, avoids these collapses and remains the strongest key-side method overall, though it is still outperformed by \ourmethod{} on sink-suppressed models.


\begin{table}[t]
\centering\small
\caption{MATH-500 reasoning retention (\% of dense flex@5 except Qwen3.5;
  pass@1 greedy; $\uparrow$; $n\!=\!100$)
  at \textbf{25\%\,/\,50\%} cache budgets, set per model as fractions of average
  sequence length (see \cref{app:math500_dense}). \textbf{Bold} marks the best
  non-dense strategy by average retention over the two budgets.
  $^\dagger$Sink on Qwen3.5-4B collapses output formatting; see
  \cref{app:sink_formatting}.}
\label{tab:math500_eviction}
\resizebox{\linewidth}{!}{%
\begin{tabular}{lcccccccccc}
\toprule
 & \multicolumn{1}{c}{\textit{Sink}} & \multicolumn{2}{c}{\textit{Attn}} & \multicolumn{3}{c}{\textit{Key Geometry}} & \multicolumn{1}{c}{\textit{Attn$\times$Val.}} & \multicolumn{3}{c}{\textit{Value Geometry}} \\
\cmidrule(lr){2-2}\cmidrule(lr){3-4}\cmidrule(lr){5-7}\cmidrule(lr){8-8}\cmidrule(lr){9-11}
Model & Streaming & TOVA & SnapKV & KeyNorm & KeyDiff & ManifoldKV & F-CAOTE & V-Norm & V-Dir & \ourmethod{} \\
\midrule
\multicolumn{11}{l}{\textit{Standard}} \\
\quad DS-R1-Qwen-1.5B
  & 74.2\,/\,89.7
  & 70.3\,/\,88.8
  & 81.1\,/\,91.1
  & 45.7\,/\,75.4
  & 89.2\,/\,99.0
  & \textbf{95.9\,/\,97.1}
  & 75.1\,/\,93.1
  & 85.4\,/\,94.5
  & 79.2\,/\,93.1
  & 89.0\,/\,97.6 \\
\quad DS-R1-Llama-8B
  & 63.1\,/\,73.9
  & 69.8\,/\,79.1
  & 76.0\,/\,83.7
  & 64.3\,/\,79.1
  & \textbf{79.4\,/\,84.9}
  & 77.0\,/\,84.9
  & 74.1\,/\,84.9
  & 77.7\,/\,83.2
  & 75.8\,/\,84.7
  & 77.9\,/\,85.1 \\
\midrule
\multicolumn{11}{l}{\textit{QK-normalization}} \\
\quad Gemma3-4B
  & 81.1\,/\,94.6
  & \textbf{94.3\,/\,104.9}
  & 88.1\,/\,98.1
  & 75.5\,/\,90.0
  & 87.1\,/\,97.8
  & 87.3\,/\,100.0
  & 95.4\,/\,103.2
  & 92.7\,/\,101.3
  & 91.9\,/\,100.5
  & 96.8\,/\,102.2 \\
\quad Qwen3-4B
  & 71.3\,/\,87.9
  & 81.7\,/\,92.0
  & 83.8\,/\,93.1
  & 67.2\,/\,86.4
  & 79.5\,/\,90.9
  & 87.3\,/\,94.2
  & 84.9\,/\,93.5
  & 87.9\,/\,95.7
  & 85.3\,/\,94.2
  & \textbf{90.5\,/\,96.3} \\
\midrule
\multicolumn{11}{l}{\textit{Gated attention (pass@1 greedy)}} \\
\quad Qwen3.5-4B
  & 1.1$^\dagger$\,/\,31.5
  & 39.3\,/\,80.9
  & 24.7\,/\,84.3
  & 40.4\,/\,75.3
  & 65.2\,/\,86.5
  & 62.9\,/\,89.9
  & 34.8\,/\,83.1
  & 76.4\,/\,94.4
  & 79.8\,/\,96.6
  & \textbf{84.3\,/\,100.0} \\
\quad Qwen3.5-9B
  & 2.2\,/\,27.0
  & 42.7\,/\,82.0
  & 19.1\,/\,73.0
  & 48.3\,/\,80.9
  & 68.5\,/\,89.9
  & 71.9\,/\,92.1
  & 49.4\,/\,89.9
  & 87.6\,/\,94.4
  & 87.6\,/\,97.8
  & \textbf{92.1\,/\,100.0} \\
\midrule
\multicolumn{11}{l}{\textit{Learned Sink}} \\
\quad GPT-OSS-20B
  & 73.9\,/\,87.1
  & 71.9\,/\,86.5
  & 74.9\,/\,88.4
  & 61.7\,/\,84.4
  & 89.1\,/\,96.1
  & 90.1\,/\,95.9
  & 75.6\,/\,90.1
  & 91.2\,/\,96.8
  & 76.7\,/\,91.0
  & \textbf{93.1\,/\,96.6} \\
\bottomrule
\end{tabular}}
\end{table}

\subsection{MATH-500}
\label{sec:math}

On MATH-500, we examine a complementary pressure point, decode-time cache
growth during chain-of-thought reasoning. We evaluate models that natively
support chain-of-thought reasoning, specifically those trained with reasoning
supervision or explicit thinking-token mechanisms~\cite{gemma2}. Budgets are
model-specific, set at approximately 25\% and 50\% of each model's average
sequence length, so compression ratios remain comparable despite different
decode lengths; dense decode lengths and truncation rates are reported in
\cref{app:math500_dense}. \ourmethod{} remains at or near the top across
budgets on sink-suppressed models and is most clearly separated on Qwen3.5
at the tight 25\% budget, where attention- and key-based
alternatives trail by roughly 19--24 points. Across sink-suppressed models,
\ourmethod{} retains at least 84\% of dense performance even at the tight
budget, showing that the value-geometry signal transfers to reasoning.

\subsection{Discussion}
\label{sec:discussion}

The empirical results above show that \ourmethod{} consistently matches or
outperforms key-geometry baselines across architectures and benchmarks.
We now characterize the geometric conditions underlying this pattern through
a lens that completes the picture established in \cref{sec:observation}.

\paragraph{BOS cosine similarity.}
Sink rate (\cref{sec:observation}) is a post-softmax measure. It summarizes the attention output after
temperature scaling and normalization, compressing the underlying key geometry
into a threshold-crossing fraction. We instead examine a more direct geometric
quantity, the cosine similarity between the BOS key vector and the mean of
content key vectors,
\begin{equation*}
\mathrm{bos\text{-}cosim} \;=\; \cos\!\bigl(k_{\mathrm{BOS}},\;
  \tfrac{1}{n-1}\textstyle\sum_{i > 0} k_i\bigr).
\end{equation*}
This quantity lives in the pre-softmax key space and is query-invariant,
measuring whether the BOS token occupies a geometrically distinctive position
relative to the content key distribution: a negative $\mathrm{bos\text{-}cosim}$
is the structural signature of attention sinks, with BOS anti-aligned with
the content mean, while a positive value indicates no such distinction.

\paragraph{Two-axis picture.}
The KeyDiff eviction score implicitly depends on this distinction.
As established theoretically in~\cite{keydiff}, KeyDiff retains tokens whose
keys diverge from the cache mean $\bar{k}$.
When $\mathrm{bos\text{-}cosim} < 0$, $\bar{k}$ is pulled toward the content
direction and BOS is a clear geometric outlier: the anchor is meaningful.
When $\mathrm{bos\text{-}cosim} > 0$, $\bar{k}$ is no longer a clean
reference and the anchor dissolves, degrading KeyDiff's discriminative signal.

\begin{figure}[t]
\centering
\includegraphics[width=\linewidth]{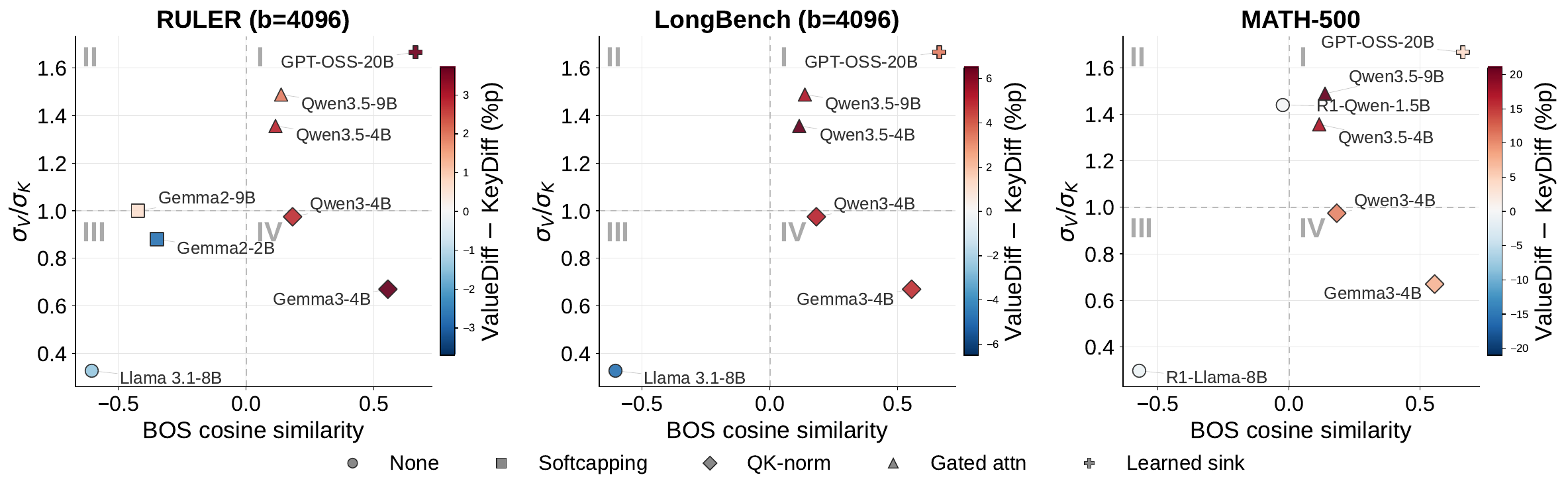}
\caption{$\mathrm{bos\text{-}cosim}$ vs.\ $\sigma_V/\sigma_K$ for all models,
  colored by \ourmethod{}$-$KeyDiff gap, across RULER (left), LongBench
  (center), and MATH-500 (right).}
\label{fig:bos_cosim_three_panel}
\vspace{-6pt}
\end{figure}

\cref{fig:bos_cosim_three_panel} plots $\mathrm{bos\text{-}cosim}$ against
$\sigma_V/\sigma_K$ for all evaluated models, colored by the
\ourmethod{}$-$KeyDiff gap, across RULER, LongBench, and MATH-500.
Two dashed lines divide the space into four quadrants.
The pattern is consistent across benchmarks.
\ourmethod{} outperforms KeyDiff in Quadrants I, II, and IV;
KeyDiff leads only in Quadrant~III ($\mathrm{bos\text{-}cosim} < 0$ and
$\sigma_V/\sigma_K < 1$), where the BOS anchor is intact and key
geometry dominates.
Gemma2-9B is particularly informative, sitting close to the $\sigma_V/\sigma_K{=}1.0$ boundary
($\mathrm{bos\text{-}cosim} \approx -0.4$,
$\sigma_V/\sigma_K \approx 1.0$) with near-parity between the two
methods (+0.6\,pt), consistent with neither signal being dominant.

\subsection{Latency and Memory}
\label{sec:latency_memory}

We focus on memory-constrained long-context inference, measuring peak memory
during long-prefill processing and time-to-first-token (TTFT). Decode
throughput depends on kernel-level cache management orthogonal to the
eviction scoring rule.

\begin{figure}[t]
\centering
\includegraphics[width=\linewidth]{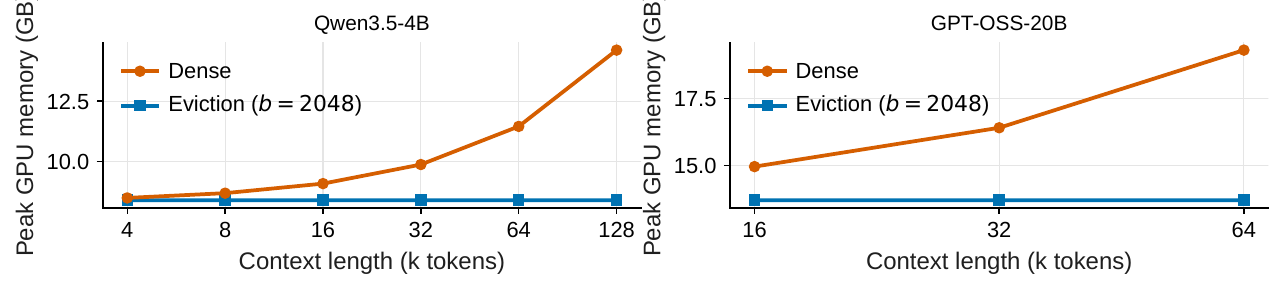}
\caption{Peak GPU memory vs.\ context length on Qwen3.5-4B and GPT-OSS-20B
  (H100 80\,GB, budget\,=\,2048). Dense scales linearly; eviction caps at
  $O(\text{budget})$. All eviction methods overlap within 0.1\%; \ourmethod{} is
  representative eviction curve.}
\label{fig:peak_memory}
\vspace{-6pt}
\end{figure}

\textbf{Peak memory.} Eviction caps the cache at $O(\text{budget})$
regardless of input length, while dense scales linearly (\cref{fig:peak_memory}).
On Qwen3.5-4B at context length 128k, peak GPU memory drops from 15.0\,GB to
8.6\,GB ($-43\%$); on GPT-OSS-20B at context length 65k the reduction is 29\%.
These savings follow from the budget itself and are shared by token eviction
methods.

\textbf{Prefill latency.} Under block prompt processing ($B{=}128$), each
chunk after cache fill attends only to the post-eviction budget-sized cache.
In our implementation, \ourmethod{} reduces TTFT relative to dense on both
long-context settings and stays within 3--4\% of the cheapest eviction
baseline. These measurements are intended to characterize scoring overhead
under a fixed block-processing implementation; optimized serving performance
also depends on kernel-level cache management and implementation details. The
full TTFT table is reported in \cref{tab:latency_memory}. Among the evaluated
eviction rules, \ourmethod{} has the same asymptotic online cost as TOVA,
SnapKV, and KeyDiff; fixed-window sink eviction is cheaper but significantly less accurate on
sink-suppressed models.

\section{Related Work}
\label{sec:related}
\paragraph{Attention sinks.}
Xiao et al.~\cite{streamingllm} identified attention sinks and proposed StreamingLLM, which preserves fixed sink tokens plus a recent window. Follow-up work examines sink emergence and control~\cite{barbero_firsttoken,sun_massive_activations,godey_anisotropy,ruscio2025sinking}. Modern architectures also include QK-normalization~\cite{qknorm}, gated
attention and hybrid linear/softmax models~\cite{gatedattn,gateddeltanet},
logit softcapping~\cite{gemma2}, and learned attention sinks~\cite{gptoss},
which are associated with weaker attention sinks. Our work connects this architectural
trend to KV eviction. As sink behavior weakens, value geometry becomes a
more useful query-invariant signal.

\paragraph{Attention-score eviction.}
H\textsubscript{2}O~\cite{h2o}, TOVA~\cite{tova}, and SnapKV~\cite{snapkv}
score cached tokens using observed attention mass or local attention windows.
KVZip~\cite{kvzip} is query-agnostic but still attention-scored. It uses a
repeat prompt to induce context reconstruction and scores KV pairs by the
maximum cross-attention they receive.
These methods rely on attention as a useful token-importance signal; our work
studies how that signal changes under modern sink-suppression mechanisms.

\paragraph{Key-geometric eviction.}
Key-geometric methods rank tokens without using attention scores. Key-norm
scoring ranks keys by $L_2$ norm~\cite{devoto_l2norm},
KeyDiff~\cite{keydiff} scores cosine distance from the mean key, and
ManifoldKV~\cite{manifoldkv} scores Euclidean distance from a key centroid,
the key-side counterpart of our L2 value score. These methods implicitly
assume key geometry is the dominant eviction signal; our two-axis analysis
(\cref{sec:discussion}) shows this is architecture-dependent: key geometry
leads where sinks are strong, value geometry leads where they weaken.
\ourmethod{} uses the value side of the cache, motivated by our observation
that value geometry becomes more discriminative than key geometry in
sink-suppressed models.

\paragraph{Attention-weighted value eviction.}
FastCAOTE~\cite{caote} is closest to \ourmethod{} in form, combining the
same centroid-distance value score with an additional query-specific attention
multiplier $\tfrac{\alpha_j}{1-\alpha_j}$. Under block prompt processing,
this multiplier is computed from past queries and may not reflect future cache
utility, introducing noise. \cref{app:proof_prop_vdiff} derives both scores
from the same reconstruction objective: \ourmethod{} arises as the consistent
max-entropy limit.

\section{Conclusion}
\label{sec:conclusion}

We revisited KV cache eviction for sink-suppressed LLMs, where models
incorporating QK-normalization, gated attention, learned attention sinks,
or logit softcapping exhibit weaker attention sinks. In this regime,
weakened sink behavior co-occurs with greater value-vector dispersion
relative to key-vector dispersion, making values a stronger
query-invariant eviction signal. This
motivates \ourmethod{}, the score
$s_j = \lVert \mathbf{v}_j - \bar{\mathbf{v}} \rVert$, which also arises as
the symmetric, max-entropy solution to minimizing attention-output disturbance
when future queries are unknown. Across RULER, LongBench, and MATH-500,
\ourmethod{} is strongest on sink-suppressed models while remaining
competitive on mixed and key-dominant models, supporting value geometry
as a more reliable eviction signal than both attention-based and
key-geometry methods in sink-suppressed architectures. The mixed-geometry
regime further motivates hybrid eviction scores that combine key- and
value-side geometry adaptively across layers as a promising direction
for future research.



\bibliographystyle{unsrt}
\bibliography{references}

\clearpage

\renewcommand \thepart{}
\renewcommand \partname{}

\newpage
\rule[0pt]{\columnwidth}{3pt}
\begin{center}
    \huge{\ourmethod{} \\
    Supplementary Material}
\end{center}
\vspace*{3mm}
\rule[0pt]{\columnwidth}{1pt}
\vspace*{-.5in}

\appendix
\addcontentsline{toc}{section}{}
\part{}
\parttoc

\renewcommand{\theequation}{A.\arabic{equation}}
\setcounter{equation}{0}

\section{ValueDiff under Dirichlet Max-Entropy, and Relationship to FastCAOTE}
\label{app:proof_prop_vdiff}

\cref{prop:vdiff_optimum} in the main body establishes \ourmethod{} as
the minimizer of the CAOTE reconstruction objective \cref{eq:recon_obj}
evaluated at the symmetric attention vector
$\boldsymbol{\alpha}^\star = (1/n, \ldots, 1/n)$, using only the
observation that $\boldsymbol{\alpha}^\star$ is the mean of any
exchangeable prior on $\Delta^{n-1}$. This appendix formalizes the
prior via Jaynes' max-entropy principle~\cite{jaynes1957}. On
$\Delta^{n-1}$ without further constraints, the maximum-entropy
distribution is $\text{Dirichlet}(1, \ldots, 1)$, a specific
exchangeable prior with strictly positive density throughout the
simplex. We (i)~recover the \ourmethod{} ranking under the full
expectation $\mathbb{E}_P[\lVert \mathbf{o}(\boldsymbol{\alpha}) - \mathbf{o}_{\setminus j}(\boldsymbol{\alpha}) \rVert^2]$
with $P = \text{Dirichlet}(1,\ldots,1)$, and (ii)~clarify the precise
relationship to FastCAOTE~\cite{caote}, whose score corresponds to a
partial max-entropy approximation of the same objective.

\paragraph{Claim.}
Let $\boldsymbol{\alpha} \sim \text{Dirichlet}(1, \ldots, 1)$
on $\Delta^{n-1}$ and $\bar{\mathbf{v}} = \tfrac{1}{n}\sum_i \mathbf{v}_i$. Then
\begin{equation}\label{eq:app_claim}
  \mathbb{E}_{\boldsymbol{\alpha}}\!\left[
    \left(\frac{\alpha_j}{1-\alpha_j}\right)^{\!2}
    \lVert \mathbf{v}_j - \mathbf{o}(\boldsymbol{\alpha}) \rVert^2
  \right]
  \;=\; \frac{2}{n(n-1)}\,\lVert \mathbf{v}_j - \bar{\mathbf{v}} \rVert^2
    \;+\; C(n),
\end{equation}
where $\mathbf{o}(\boldsymbol{\alpha}) = \sum_i \alpha_i \mathbf{v}_i$ and
$C(n)$ depends only on $n$ and the cache (not on $j$). The integrand is
the per-query squared reconstruction error after evicting $j$ and
renormalizing the softmax, with
$\lVert \mathbf{o} - \mathbf{o}' \rVert^2 =
(\alpha_j/(1-\alpha_j))^2 \lVert \mathbf{v}_j - \mathbf{o} \rVert^2$
and $\mathbf{o}' = (\mathbf{o} - \alpha_j \mathbf{v}_j)/(1 - \alpha_j)$.
The minimizer is
$j^\star = \arg\min_j \lVert \mathbf{v}_j - \bar{\mathbf{v}} \rVert$,
matching the main-body proposition and confirming that \ourmethod{} is
robust to the specific form of max-entropy adopted.

The proof proceeds in three steps. First, a renormalization identity rewrites
the integrand in a more tractable form. Second, a Dirichlet independence
property factors the expectation. Third, elementary moment calculations finish
the reduction.

\paragraph{Step 1. Renormalization identity.}
Define the post-eviction output
$\mathbf{o}'_j := \sum_{i \neq j} (\alpha_i / (1-\alpha_j)) \mathbf{v}_i$.
Direct algebra gives
\begin{equation*}
  \mathbf{v}_j - \mathbf{o}
    \;=\; (1 - \alpha_j)\mathbf{v}_j - \sum_{i \neq j} \alpha_i \mathbf{v}_i
    \;=\; (1 - \alpha_j)(\mathbf{v}_j - \mathbf{o}'_j),
\end{equation*}
so that
\begin{equation}\label{eq:app_renorm_identity}
  \left(\frac{\alpha_j}{1-\alpha_j}\right)^2 \lVert \mathbf{v}_j - \mathbf{o} \rVert^2
  \;=\; \alpha_j^2 \lVert \mathbf{v}_j - \mathbf{o}'_j \rVert^2.
\end{equation}

\paragraph{Step 2. Dirichlet independence.}
A standard property of the Dirichlet distribution states that if
$\boldsymbol{\alpha} \sim \text{Dirichlet}(a_1, \ldots, a_n)$, then the
marginal $\alpha_j \sim \text{Beta}(a_j, \sum_{i \neq j} a_i)$ is
\emph{independent} of the renormalized other weights
$\boldsymbol{\beta} = (\alpha_i / (1 - \alpha_j))_{i \neq j} \sim \text{Dirichlet}(a_{-j})$.
In the symmetric case $a_i = 1$ this gives
\begin{equation*}
  \alpha_j \sim \text{Beta}(1, n-1),
  \qquad
  \boldsymbol{\beta} \sim \text{Dirichlet}(1, \ldots, 1)
    \text{ on } \Delta^{n-2},
  \qquad
  \alpha_j \perp \boldsymbol{\beta}.
\end{equation*}
Since $\mathbf{o}'_j$ is a function only of $\boldsymbol{\beta}$ (and the
fixed cache values), $\mathbf{o}'_j \perp \alpha_j$. Combined with
\cref{eq:app_renorm_identity},
\begin{equation}\label{eq:app_factored}
  \mathbb{E}\!\left[ \left(\frac{\alpha_j}{1-\alpha_j}\right)^2
      \lVert \mathbf{v}_j - \mathbf{o} \rVert^2 \right]
  \;=\; \mathbb{E}[\alpha_j^2] \cdot
        \mathbb{E}[\lVert \mathbf{v}_j - \mathbf{o}'_j \rVert^2].
\end{equation}

\paragraph{Step 3. Moments.}
\emph{First factor.} For $\alpha_j \sim \text{Beta}(1, n-1)$, standard Beta
moments give $\mathbb{E}[\alpha_j] = 1/n$ and
\begin{equation}\label{eq:app_first_moment}
  \mathbb{E}[\alpha_j^2] \;=\; \frac{2}{n(n+1)}.
\end{equation}

\emph{Second factor.} Since $\sum_{i \neq j} \beta_i = 1$, we can rewrite
\begin{equation}\label{eq:app_vo_diff}
  \mathbf{v}_j - \mathbf{o}'_j
    \;=\; \sum_{i \neq j} \beta_i (\mathbf{v}_j - \mathbf{v}_i)
    \;=\; \sum_{i \neq j} \beta_i\, \mathbf{d}_i,
\end{equation}
where $\mathbf{d}_i := \mathbf{v}_j - \mathbf{v}_i$. Standard Dirichlet(1,\ldots,1)
moments on $\Delta^{n-2}$ give $\mathbb{E}[\beta_i^2] = 2/((n-1)n)$ and
$\mathbb{E}[\beta_i \beta_k] = 1/((n-1)n)$ for $i \neq k$, which can be
combined as $\mathbb{E}[\beta_i \beta_k] = (1 + \mathbf{1}[i=k])/((n-1)n)$.
Therefore
\begin{equation}\label{eq:app_dirichlet_quad}
  \mathbb{E}\!\left[\bigl\lVert \textstyle\sum_{i \neq j} \beta_i \mathbf{d}_i \bigr\rVert^2\right]
  \;=\; \frac{1}{(n-1)n}\!\left(\textstyle\sum_{i \neq j} \lVert \mathbf{d}_i \rVert^2
        \;+\; \bigl\lVert \sum_{i \neq j} \mathbf{d}_i \bigr\rVert^2\right).
\end{equation}

Both sums reduce to cache-level quantities. Using
$\sum_{i \neq j} \mathbf{v}_i = n\bar{\mathbf{v}} - \mathbf{v}_j$,
\begin{equation}\label{eq:app_sum_d}
  \Bigl\lVert \textstyle\sum_{i \neq j} \mathbf{d}_i \Bigr\rVert^2
    \;=\; \bigl\lVert (n-1)\mathbf{v}_j - (n\bar{\mathbf{v}} - \mathbf{v}_j) \bigr\rVert^2
    \;=\; n^2 \lVert \mathbf{v}_j - \bar{\mathbf{v}} \rVert^2.
\end{equation}
Expanding the squared distances with $S_2 := \sum_i \lVert \mathbf{v}_i \rVert^2$
and using $\lVert \mathbf{v}_j \rVert^2 - 2\mathbf{v}_j^\top \bar{\mathbf{v}}
= \lVert \mathbf{v}_j - \bar{\mathbf{v}} \rVert^2 - \lVert \bar{\mathbf{v}} \rVert^2$,
\begin{equation}\label{eq:app_sum_d2}
  \sum_{i \neq j} \lVert \mathbf{d}_i \rVert^2
    \;=\; n\lVert \mathbf{v}_j \rVert^2 - 2n\mathbf{v}_j^\top \bar{\mathbf{v}} + S_2
    \;=\; n\lVert \mathbf{v}_j - \bar{\mathbf{v}} \rVert^2
        + \bigl(S_2 - n\lVert \bar{\mathbf{v}} \rVert^2\bigr).
\end{equation}

Substituting \cref{eq:app_sum_d,eq:app_sum_d2} into \cref{eq:app_dirichlet_quad},
\begin{equation}\label{eq:app_second_factor}
  \mathbb{E}[\lVert \mathbf{v}_j - \mathbf{o}'_j \rVert^2]
  \;=\; \frac{n+1}{n-1}\,\lVert \mathbf{v}_j - \bar{\mathbf{v}} \rVert^2
        \;+\; \frac{S_2 - n\lVert \bar{\mathbf{v}} \rVert^2}{(n-1)n}.
\end{equation}
The second term depends only on $n$ and the cache, not on $j$.

\paragraph{Conclusion.}
Substituting \cref{eq:app_first_moment,eq:app_second_factor} into \cref{eq:app_factored},
\begin{align*}
  \mathbb{E}\!\left[ \left(\frac{\alpha_j}{1-\alpha_j}\right)^2
      \lVert \mathbf{v}_j - \mathbf{o} \rVert^2 \right]
  &\;=\; \frac{2}{n(n+1)} \cdot \left(
       \frac{n+1}{n-1} \lVert \mathbf{v}_j - \bar{\mathbf{v}} \rVert^2
     + \frac{S_2 - n\lVert \bar{\mathbf{v}} \rVert^2}{(n-1)n} \right) \\
  &\;=\; \frac{2}{n(n-1)} \lVert \mathbf{v}_j - \bar{\mathbf{v}} \rVert^2
       + \underbrace{\frac{2\,(S_2 - n \lVert \bar{\mathbf{v}} \rVert^2)}{n^2(n-1)(n+1)}}_{=:\, C(n)},
\end{align*}
which is \cref{eq:app_claim}. Since $C(n)$ does not depend on $j$, the
minimizer of the expected reconstruction error is
$j^\star = \arg\min_j \lVert \mathbf{v}_j - \bar{\mathbf{v}} \rVert$. \hfill$\blacksquare$

\paragraph{Relationship to FastCAOTE.}
The renormalization identity in Step~1, taken in unsquared form, reads
\begin{equation}\label{eq:app_unsquared}
  \bigl\lVert \mathbf{o}(\boldsymbol{\alpha}) - \mathbf{o}'_j \bigr\rVert
  \;=\; \frac{\alpha_j}{1-\alpha_j}\, \lVert \mathbf{v}_j - \mathbf{o}(\boldsymbol{\alpha}) \rVert,
\end{equation}
where $\mathbf{o}'_j = \sum_{i\neq j}(\alpha_i/(1-\alpha_j))\mathbf{v}_i$ is
the post-eviction softmax-renormalized output. FastCAOTE~\cite{caote}
evaluates \cref{eq:app_unsquared} at an observed attention vector
$\tilde{\boldsymbol{\alpha}}$ and applies a single approximation,
$\mathbf{o}(\tilde{\boldsymbol{\alpha}}) \to \bar{\mathbf{v}}$, yielding the
score
\begin{equation}\label{eq:app_fastcaote}
  s_j^{\mathrm{FastCAOTE}}
  \;=\; \frac{\tilde\alpha_j}{1-\tilde\alpha_j}\, \lVert \mathbf{v}_j - \bar{\mathbf{v}} \rVert
\end{equation}
(Eq.~23 of~\cite{caote}). The substitution
$\mathbf{o} \to \bar{\mathbf{v}}$ is itself max-entropy in character. Under
$\boldsymbol{\alpha} \sim \text{Uniform}$,
$\mathbb{E}[\mathbf{o}(\boldsymbol{\alpha})] = \bar{\mathbf{v}}$, and the
approximation is exact at $\boldsymbol{\alpha}^\star$. The factor
$\tilde\alpha_j/(1-\tilde\alpha_j)$, in contrast, retains the observed
attention weight without a corresponding max-entropy substitution.

Applying max-entropy consistently to \emph{both} quantities, as \ourmethod{}
does, additionally substitutes $\alpha_j \to 1/n$, giving
$\alpha_j/(1-\alpha_j) \to 1/(n-1)$. This collapses \cref{eq:app_fastcaote}
to $\lVert \mathbf{v}_j - \bar{\mathbf{v}} \rVert / (n-1)$, and the
$j$-independent factor $1/(n-1)$ drops out of the argmin, recovering
\ourmethod{}. From this angle, \ourmethod{} is the consistent max-entropy
limit of the same reconstruction error in which FastCAOTE is rooted,
differing only in that the attention weighting is also max-entropy
approximated and thus vanishes from the ranking.

\section{Distance from the Symmetric Limit}
\label{app:attention_flatness}

The symmetric reference point in \cref{sec:theory} is useful because it makes
the absence of future query information explicit. In this appendix we analyze
the eviction error in the more general setting where $\boldsymbol{\alpha}$
may deviate from $\boldsymbol{\alpha}^*$, and quantify how that deviation
controls the gap from the exact rule.
For any realized attention
vector, the attention output decomposes as
\begin{equation}\label{eq:app_output_decomp}
  \mathbf{o}(\boldsymbol{\alpha})
  =
  \sum_i \alpha_i \mathbf{v}_i
  =
  \bar{\mathbf{v}}
  +
  \sum_i\left(\alpha_i-\tfrac{1}{n}\right)\mathbf{v}_i ,
  \qquad
  \bar{\mathbf{v}}=\tfrac{1}{n}\sum_i\mathbf{v}_i .
\end{equation}
Substituting \cref{eq:app_output_decomp} into \cref{eq:app_unsquared}
makes the dependence on the attention weights explicit:
\begin{equation}\label{eq:app_eviction_decomp}
  e_j(\boldsymbol{\alpha})
  = \frac{\alpha_j}{1-\alpha_j}
    \Bigl\|(\mathbf{v}_j - \bar{\mathbf{v}})
           - \textstyle\sum_i\bigl(\alpha_i - \tfrac{1}{n}\bigr)\mathbf{v}_i\Bigr\|.
\end{equation}
When $\boldsymbol{\alpha} = \boldsymbol{\alpha}^*$, the sum
$\sum_i(\alpha_i-1/n)\mathbf{v}_i$ vanishes and $\alpha_j = 1/n$, so
\cref{eq:app_eviction_decomp} reduces to the \ourmethod{} score
$s_j = \lVert\mathbf{v}_j - \bar{\mathbf{v}}\rVert$ scaled by $1/(n-1)$,
recovering
\cref{prop:vdiff_optimum} exactly. To bound \cref{eq:app_eviction_decomp}
for general $\boldsymbol{\alpha}$, we introduce a scalar measuring how far
$\boldsymbol{\alpha}$ departs from $\boldsymbol{\alpha}^*$:
\begin{equation}\label{eq:app_delta}
  \delta(\boldsymbol{\alpha})
  =
  \left\|\boldsymbol{\alpha} - \tfrac{1}{n}\mathbf{1}\right\|_1,
\end{equation}
the total attention mass displaced from uniform --- zero when
$\boldsymbol{\alpha} = \boldsymbol{\alpha}^*$ and growing as attention
concentrates. $\delta(\boldsymbol{\alpha})$ captures the attention weight departure, but the
resulting displacement of the attention output also depends on how spread out
the cached value vectors are. The following bound makes this connection
explicit, combining the attention deviation $\delta$ with the value geometry
to translate one into an output error.

For the sum in \cref{eq:app_eviction_decomp}, let $\varepsilon_i = \alpha_i - 1/n$ and let
$\mathcal{V}=\{\mathbf{v}_i\}_{i=1}^n$,
$\mathrm{diam}(\mathcal{V})=\max_{i,k}\lVert\mathbf{v}_i-\mathbf{v}_k\rVert$.
Since $\sum_i\varepsilon_i=0$, the total positive mass
$D^+=\sum_{i:\varepsilon_i>0}\varepsilon_i$ equals the total negative mass
$\sum_{i:\varepsilon_i<0}|\varepsilon_i|$, and both equal $\delta/2$.
Define convex combinations $\mathbf{p}=\sum_{i:\varepsilon_i>0}(\varepsilon_i/D^+)\mathbf{v}_i$
and $\mathbf{q}=\sum_{i:\varepsilon_i<0}(|\varepsilon_i|/D^-)\mathbf{v}_i$;
then $\sum_i\varepsilon_i\mathbf{v}_i = \tfrac{\delta}{2}(\mathbf{p}-\mathbf{q})$
and $\lVert\mathbf{p}-\mathbf{q}\rVert\le\mathrm{diam}(\mathcal{V})$.
Therefore
\begin{equation}\label{eq:app_delta_output}
  \left\|\sum_i \left(\alpha_i-\tfrac{1}{n}\right)\mathbf{v}_i\right\|
  \le
  \frac{\delta(\boldsymbol{\alpha})}{2}\,\mathrm{diam}(\mathcal{V}).
\end{equation}
Applying the triangle inequality to \cref{eq:app_eviction_decomp} with
\cref{eq:app_delta_output} and $\alpha_j \le 1/n + \delta(\boldsymbol{\alpha})$,
\begin{equation}\label{eq:app_eviction_bound}
  e_j(\boldsymbol{\alpha}) \;\leq\;
  \underbrace{\varphi\bigl(\delta(\boldsymbol{\alpha})\bigr)}_{\text{attention factor}}
  \Bigl(
  \underbrace{s_j}_{\text{\ourmethod{} score}}
  \;+\;
  \tfrac{\delta(\boldsymbol{\alpha})}{2}\,\underbrace{\mathrm{diam}(\mathcal{V})}_{\text{value spread}}
  \Bigr),
  \qquad
  \varphi(\delta) \;=\; \frac{\tfrac{1}{n}+\delta}{1-\tfrac{1}{n}-\delta}.
\end{equation}
The attention factor $\varphi(\delta)$ captures how far $\alpha_j$ departs
from $1/(n-1)$ due to non-uniform attention. The \ourmethod{} score $s_j$
is the exact eviction cost at the symmetric point.

\section{KV Geometry Across Architectures}
\label{app:per_layer_geometry}

\subsection*{Aggregate statistics}

\cref{tab:kv_diversity_full} reports sink rate and key/value L2 diversity
for all 11 models in the same order as \cref{tab:ruler_main}. Statistics are
computed on 20 calibration prompts spanning diverse content categories
(narrative, QA, code, math, retrieval), each processed at context length 2048. $\sigma_K$ and $\sigma_V$ are the standard deviation of per-token L2 distances from the cache mean, averaged over layers, heads, and the 20 prompts.

\begin{table}[H]
\centering
\small
\caption{Sink rate and K/V L2 diversity across 11 models. $\sigma_V/\sigma_K > 1$ indicates value-dominant geometry.}
\label{tab:kv_diversity_full}
\begin{tabular}{llcccc}
\toprule
Model & Mechanism & Sink rate & $\sigma_K$ & $\sigma_V$ & $\sigma_V/\sigma_K$ \\
\midrule
Llama 3.1-8B    & None          & 0.801 & 1.783 & 0.583 & 0.33 \\
Llama 3.2-3B    & None          & 0.835 & 1.478 & 0.584 & 0.40 \\
Qwen2.5-3B      & None          & 0.446 & 1.561 & 1.851 & 1.19 \\
Qwen2.5-7B      & None          & 0.392 & 1.597 & 2.133 & 1.34 \\
\midrule
Gemma2-2B       & Softcapping   & 0.464 & 1.807 & 1.590 & 0.88 \\
Gemma2-9B       & Softcapping   & 0.542 & 1.808 & 1.808 & 1.00 \\
Qwen3-4B        & QK-norm       & 0.548 & 2.322 & 2.263 & 0.97 \\
Gemma3-4B       & QK-norm       & 0.271 & 12.506 & 8.377 & 0.67 \\
\midrule
Qwen3.5-4B      & Gated attn    & 0.015 & 1.596 & 2.164 & 1.36 \\
Qwen3.5-9B      & Gated attn    & 0.012 & 2.344 & 3.488 & 1.49 \\
GPT-OSS-20B     & Learned sink  & 0.029 & 2.191 & 3.652 & 1.67 \\
\bottomrule
\end{tabular}
\end{table}

\subsection*{BOS token Q-K alignment and key-space geometric distinctiveness}

\cref{tab:sink_dual_mechanism} reports two complementary signals that
characterize the attention sink mechanism across our 11 evaluation models.
\emph{BOS Q-K weight} ($\bar{\alpha}_{\mathrm{BOS}}$) is the mean attention
mass received by the BOS token from non-BOS query positions ($t \geq 1$),
\begin{equation*}
  \bar{\alpha}_{\mathrm{BOS}}
  \;=\;
  \frac{1}{T-1}\sum_{t=1}^{T-1} \alpha_{t,0},
\end{equation*}
where $\alpha_{t,0}$ is the attention weight query position $t$ assigns to
position~0 (BOS), and the sum excludes $t{=}0$ since the BOS token can only
attend to itself under causal masking. High values indicate the BOS key
vector aligns strongly with most query vectors, yielding disproportionate
attention regardless of query content.
\emph{BOS key cosim} is the cosine similarity between the BOS key vector
and the mean of non-BOS key vectors ($i \geq 1$),
\begin{equation*}
  \mathrm{bos\text{-}cossim}
  \;=\;
  \cos\!\Bigl(k_0,\;\tfrac{1}{T-1}\textstyle\sum_{i=1}^{T-1} k_i\Bigr),
\end{equation*}
where $k_0$ is the BOS key and $k_i$ ($i \geq 1$) are the remaining key vectors. Negative
values indicate the BOS key is a geometric outlier in key space
(\cref{fig:bos_cosim_per_layer}).
Together, these two signals capture the dual mechanism that prior eviction
methods rely on: high Q-K alignment makes BOS a reliable attentional anchor,
while geometric distinctiveness from content keys makes BOS a reliable
key-geometry outlier. Both signals weaken in architectures where sinks are
suppressed.

\begin{table}[h]
\centering\small
\caption{BOS token Q-K alignment and key-space geometric distinctiveness across
  11 models. High $\bar{\alpha}_{\mathrm{BOS}}$ indicates strong persistent
  Q-K alignment (attentional anchor). Negative BOS key cosim indicates
  the BOS key is anti-aligned with the content key mean (geometric outlier).}
\label{tab:sink_dual_mechanism}
\begin{tabular}{llcc}
\toprule
Model & Mechanism & $\bar{\alpha}_{\mathrm{BOS}}$ & BOS key cos. sim. \\
\midrule
Llama~3.1-8B  & None         & 0.591 & $-$0.604 \\
Llama~3.2-3B  & None         & 0.612 & $-$0.586 \\
Qwen2.5-3B    & None         & 0.317 & \phantom{$-$}0.091 \\
Qwen2.5-7B    & None         & 0.279 & \phantom{$-$}0.337 \\
\midrule
Gemma2-2B     & Softcapping  & 0.333 & $-$0.350 \\
Gemma2-9B     & Softcapping  & 0.396 & $-$0.424 \\
Qwen3-4B      & QK-norm      & 0.408 & \phantom{$-$}0.182 \\
Gemma3-4B     & QK-norm      & 0.199 & \phantom{$-$}0.555 \\
\midrule
Qwen3.5-4B    & Gated attn   & 0.034 & \phantom{$-$}0.115 \\
Qwen3.5-9B    & Gated attn   & 0.031 & \phantom{$-$}0.137 \\
GPT-OSS-20B   & Learned sink & 0.030 & \phantom{$-$}0.663 \\
\bottomrule
\end{tabular}
\end{table}

To further characterize the geometric regime difference, we compare
$\sigma_V/\sigma_K$ against two additional metrics.
The \emph{weight-space spectral norm ratio}
$\lVert\mathbf{W}_V\rVert_2 / \lVert\mathbf{W}_K\rVert_2$ is entirely
data-independent: it requires no calibration text and captures only how
much each projection amplifies a unit-norm input, a property fixed in the
model weights.
The \emph{activation-space spectral norm ratio}
$\sigma_1(V) / \sigma_1(K)$, where $\sigma_1(\cdot)$ denotes the largest
singular value of the stacked activation matrix, is data-dependent like
$\sigma_V/\sigma_K$ but measures a different geometric property
(dominant variance direction rather than radial spread).

\begin{table}[h]
\centering\small
\caption{Three geometric metrics across architectures. The weight ratio
  is data-independent (projection matrix structure only); the other two
  require forward passes on calibration text. All three preserve the same
  architectural ordering.}
\label{tab:spectral_norm}
\begin{tabular}{llccc}
\toprule
Model & Mechanism
  & $\lVert W_V\rVert_2 / \lVert W_K\rVert_2$
  & $\sigma_1(V)/\sigma_1(K)$
  & $\sigma_V/\sigma_K$ \\
\midrule
Llama~3.1-8B  & None         & 0.192 & 0.057 & 0.33 \\
Llama~3.2-3B  & None         & 0.195 & 0.071 & 0.40 \\
Qwen2.5-3B    & None         & 0.668 & 0.216 & 1.19 \\
Qwen2.5-7B    & None         & 0.743 & 0.194 & 1.34 \\
\midrule
Gemma2-2B     & Softcapping  & 0.550 & 0.257 & 0.88 \\
Gemma2-9B     & Softcapping  & 0.632 & 0.288 & 1.00 \\
Qwen3-4B      & QK-norm      & 0.651 & 0.320 & 1.00 \\
Gemma3-4B     & QK-norm      & 0.497 & 0.397 & 0.67 \\
\midrule
Qwen3.5-4B    & Gated attn   & 1.174 & 1.358 & 1.36 \\
Qwen3.5-9B    & Gated attn   & 1.134 & 1.339 & 1.49 \\
GPT-OSS-20B   & Learned sink & 1.120 & 0.815 & 1.67 \\
\bottomrule
\end{tabular}
\end{table}

Two observations follow from \cref{tab:spectral_norm}. First, all three
metrics preserve the same architectural ordering: Llama at the
key-dominant extreme, Qwen3.5 and GPT-OSS at the value-dominant extreme,
with intermediate architectures between. Second, and more importantly,
the weight ratio and the activation ratio diverge substantially in
magnitude. For Llama,
$\lVert\mathbf{W}_V\rVert_2/\lVert\mathbf{W}_K\rVert_2 = 0.19$ but
$\sigma_1(V)/\sigma_1(K) = 0.06$---the activation-level key dominance is
roughly three times stronger than the weight-level bias alone predicts.
For Qwen3.5 the gap runs in the opposite direction ($1.17$ vs $1.36$).
This systematic divergence shows that token content contributes an
independent geometric signal on top of the projection structure:
$\sigma_V/\sigma_K$ measures a genuine property of how tokens distribute
in the projected spaces, not an artifact of how the weight matrices are
scaled.

\subsection*{Per-layer sink rate}

\begin{figure}[H]
\centering
\includegraphics[width=\linewidth]{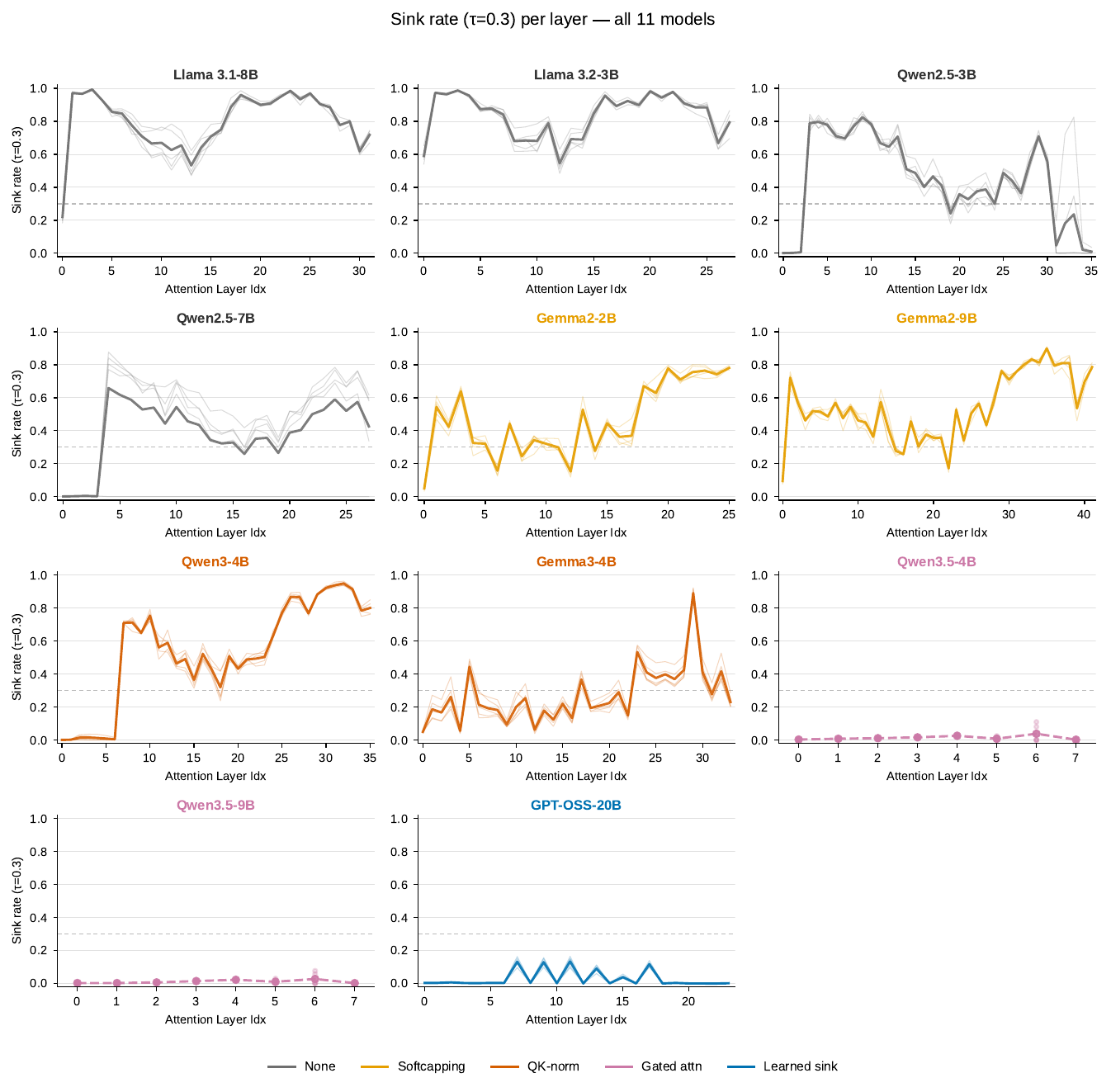}
\caption{Per-layer sink rate ($\tau = 0.3$) for all 11 models.
  Llama maintains high sink rates throughout; Qwen3.5 and GPT-OSS show
  near-zero; mixed-regime models show intermediate rates.
  Qwen2.5-3B uniquely exhibits sharp oscillations attributable to its 2 KV heads.}
\label{fig:sink_rate_per_layer}
\end{figure}

\cref{fig:sink_rate_per_layer} shows the per-layer sink rate ($\tau = 0.3$)
for all 11 models. Non-suppressed models (Llama) maintain high sink rates
consistently across all layers. Sink-suppression models split into two patterns:
gated-attention and learned-bias models (Qwen3.5, GPT-OSS) show near-zero sink
rates throughout, while QK-norm and softcapping models (Qwen3, Gemma2, Gemma3)
show intermediate rates with moderate layer-to-layer variation.
Qwen2.5 is the sole exception with sharp, alternating oscillations discussed
further in \cref{subsec:qwen25_anomaly}.

\subsection*{Per-layer $\sigma_K$ and $\sigma_V$}

\begin{figure}[H]
\centering
\includegraphics[width=\linewidth]{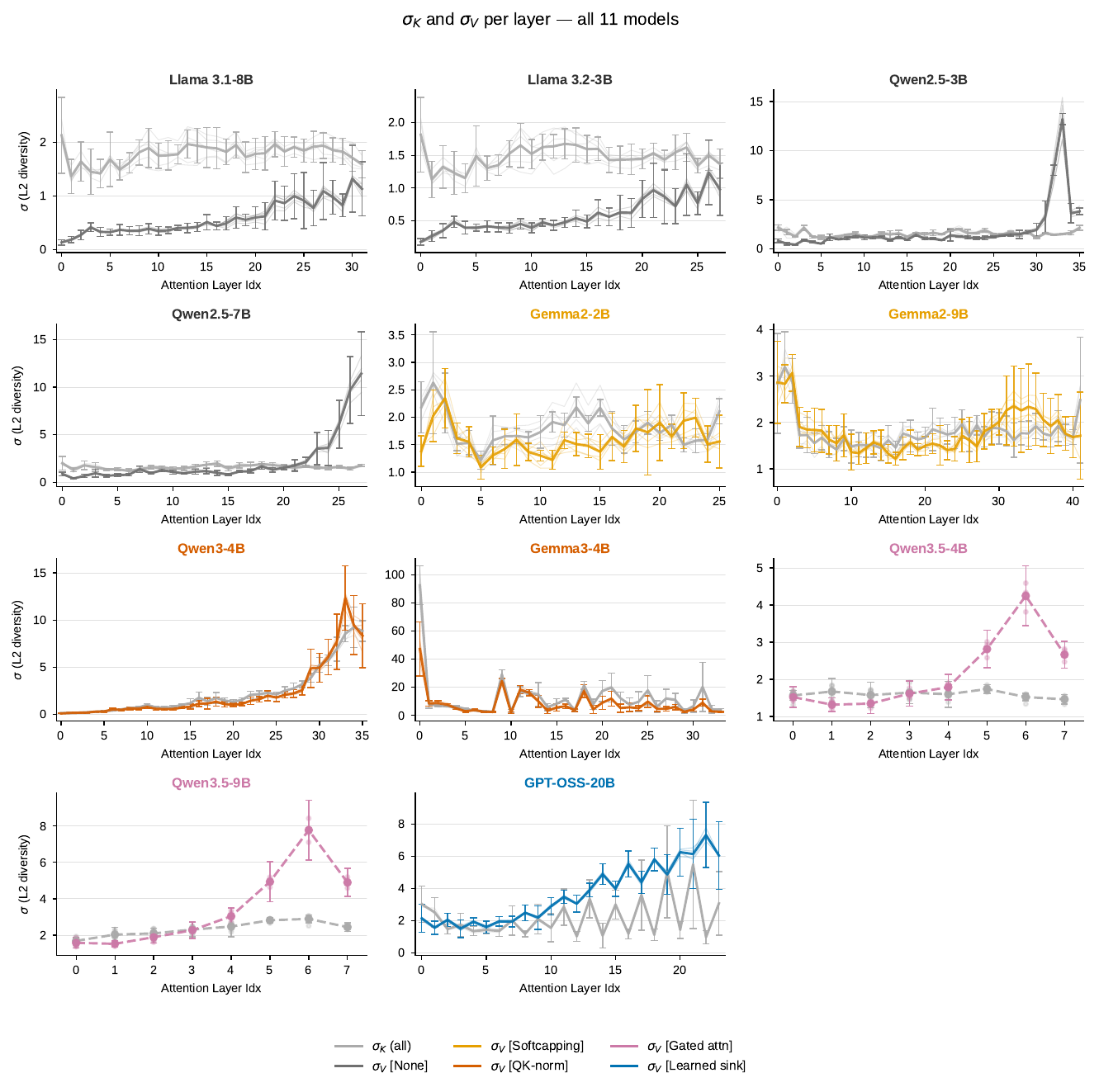}
\caption{Per-layer key and value L2 diversity ($\sigma_K^{(\ell)}$, grey;
  $\sigma_V^{(\ell)}$, mechanism color) for all 11 models. Error bars show
  per-head standard deviation. Qwen3.5 shows only 8 full-attention layers.}
\label{fig:sigma_kv_per_layer}
\end{figure}

\cref{fig:sigma_kv_per_layer} shows $\sigma_K^{(\ell)}$ and $\sigma_V^{(\ell)}$
at each attention layer for all 11 models. Error bars reflect per-head standard
deviation within each layer, averaged over samples. For Qwen3.5, only the 8
full-attention layers are shown; the x-axis indexes attention layers sequentially.

\subsection*{Per-layer BOS key cosine similarity}

\begin{figure}[H]
\centering
\includegraphics[width=\linewidth]{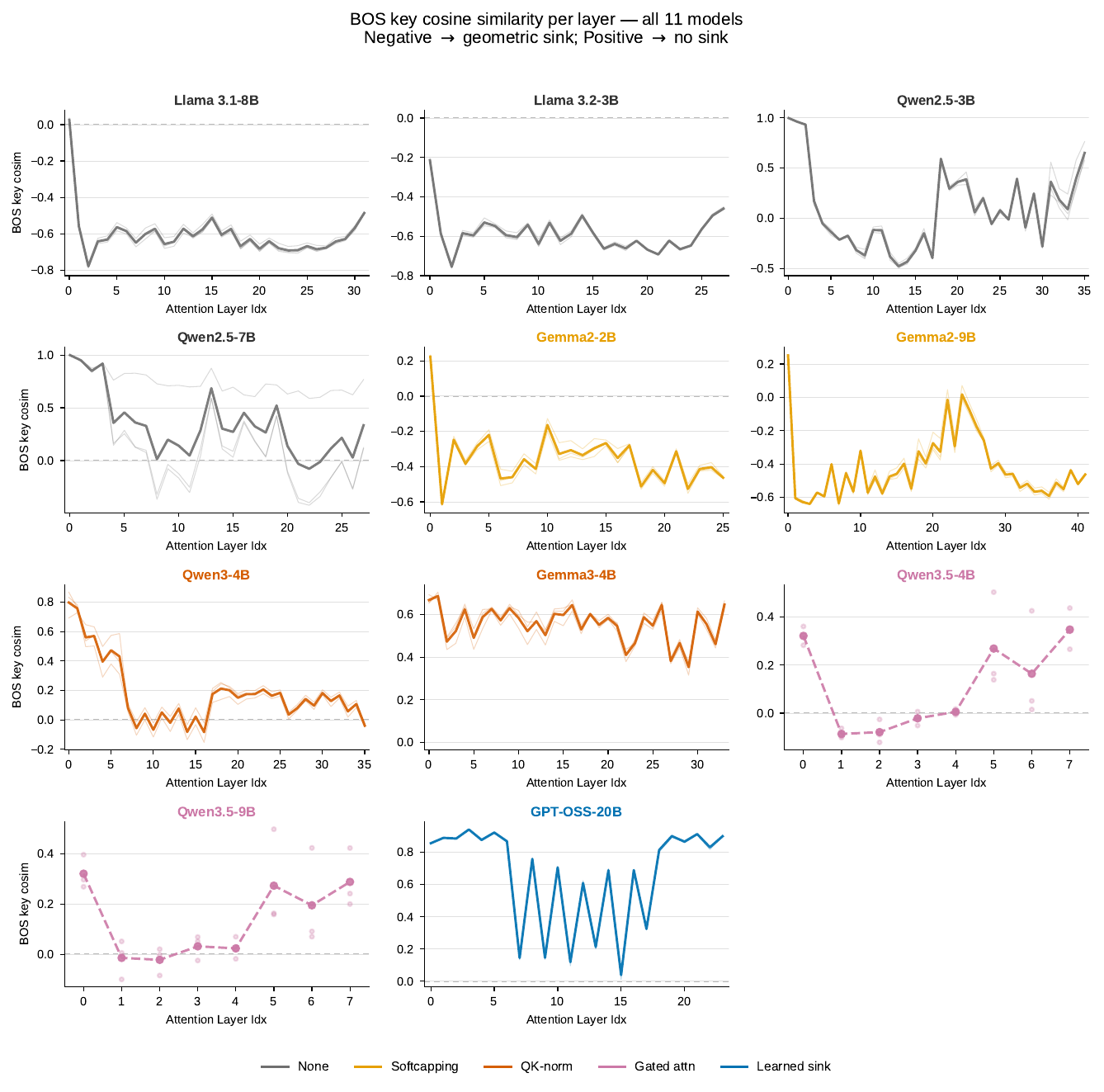}
\caption{Per-layer BOS key cosine similarity for all 11 models.
  Values below zero indicate a geometrically distinct BOS key (attention sink);
  values above zero indicate sink suppression.
  The dashed line marks $\mathrm{bos\text{-}cossim} = 0$.}
\label{fig:bos_cosim_per_layer}
\end{figure}

\cref{fig:bos_cosim_per_layer} shows the mean BOS key cosine similarity
$\mathrm{bos\text{-}cossim}^{(\ell)}$ at each attention layer.
Negative values indicate the BOS key vector is a geometric outlier
relative to the content key mean---the structural signature of an attention
sink. Positive values indicate BOS is not geometrically distinct from
content keys. The pattern mirrors the aggregate picture in
\cref{fig:sink_vs_ratio_polished}: Llama maintains consistently negative
values throughout all layers (strong geometric sink), Qwen3.5 and GPT-OSS
are near zero or positive (sink suppressed), and QK-norm and softcapping
models (Qwen3, Gemma2, Gemma3) sit in between.

\subsection*{Per-layer $\sigma_V/\sigma_K$}

\begin{figure}[H]
\centering
\includegraphics[width=\linewidth]{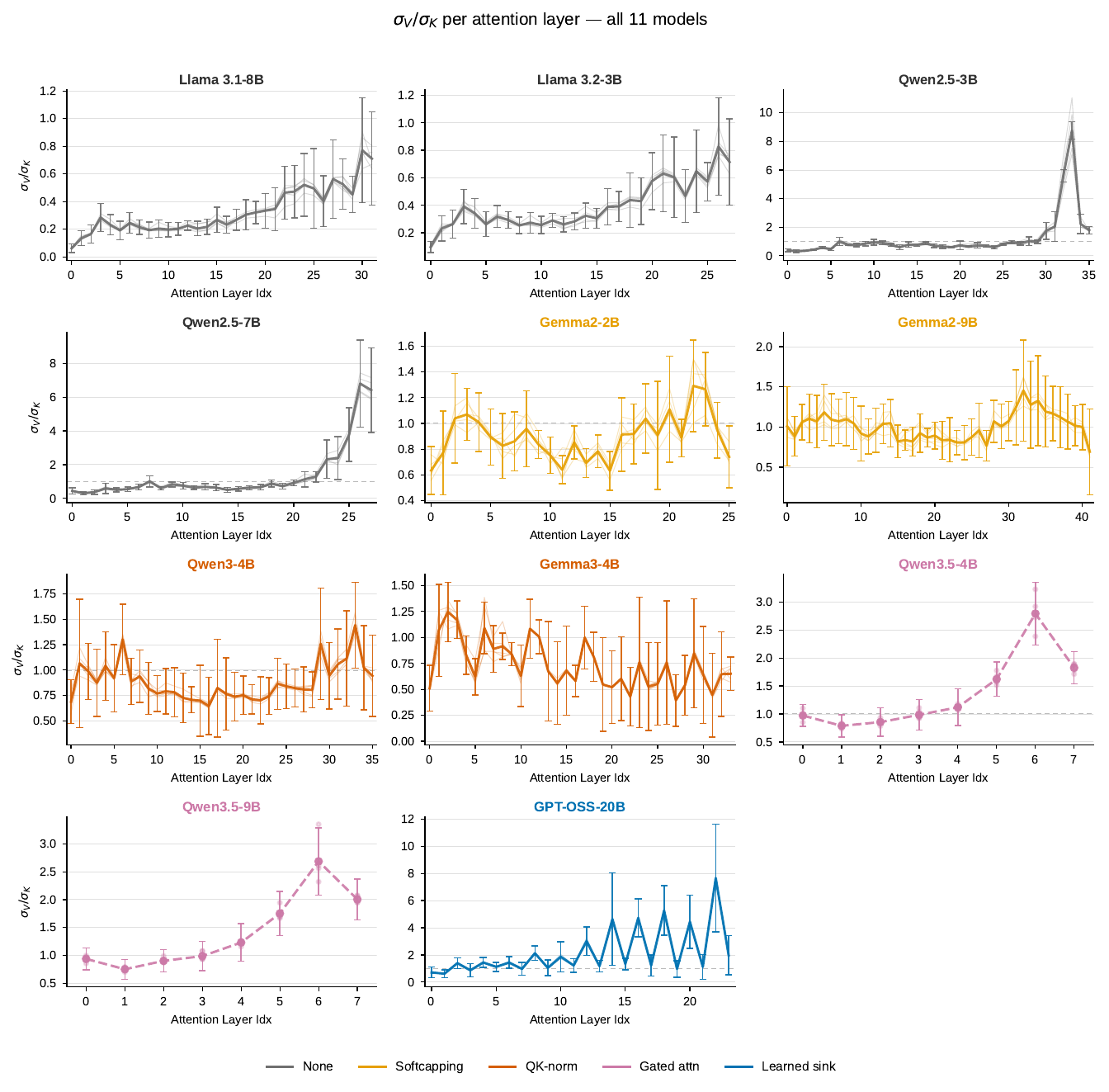}
\caption{Per-layer $\sigma_V/\sigma_K$ ratio for all 11 models. The dashed
  horizontal line marks $\sigma_V = \sigma_K$. All models except Qwen2.5
  maintain a coherent regime across layers.}
\label{fig:sigma_ratio_per_layer}
\end{figure}

\cref{fig:sigma_ratio_per_layer} shows the per-layer $\sigma_V/\sigma_K$ ratio. The per-layer profiles
reflect the three geometry regimes discussed in \cref{sec:ruler_results}.
\textbf{Value-dominant models} (GPT-OSS-20B, Qwen3.5) show consistently high
$\sigma_V$, either uniform throughout or strengthening in later full-attention
layers, providing a stable value-side signal that \ourmethod{} exploits.
\textbf{Mixed-regime models} (Qwen3, Gemma2, Gemma3) oscillate around
$\sigma_V/\sigma_K \approx 1$ within a moderate range (0.6--1.6); even here,
\ourmethod{} generally comes out ahead as the value signal retains an edge
(\cref{sec:ruler_results}). \textbf{Key-dominant models} (Llama) are
consistently below 1 across all layers, where KeyDiff is the strongest method
and \ourmethod{} remains competitive. Each of these patterns is geometrically
coherent, and a single eviction strategy applies consistently within each model.

\subsection*{The Qwen2.5 Anomaly}
\label{subsec:qwen25_anomaly}

\begin{figure}[H]
\centering
\includegraphics[width=0.80\linewidth]{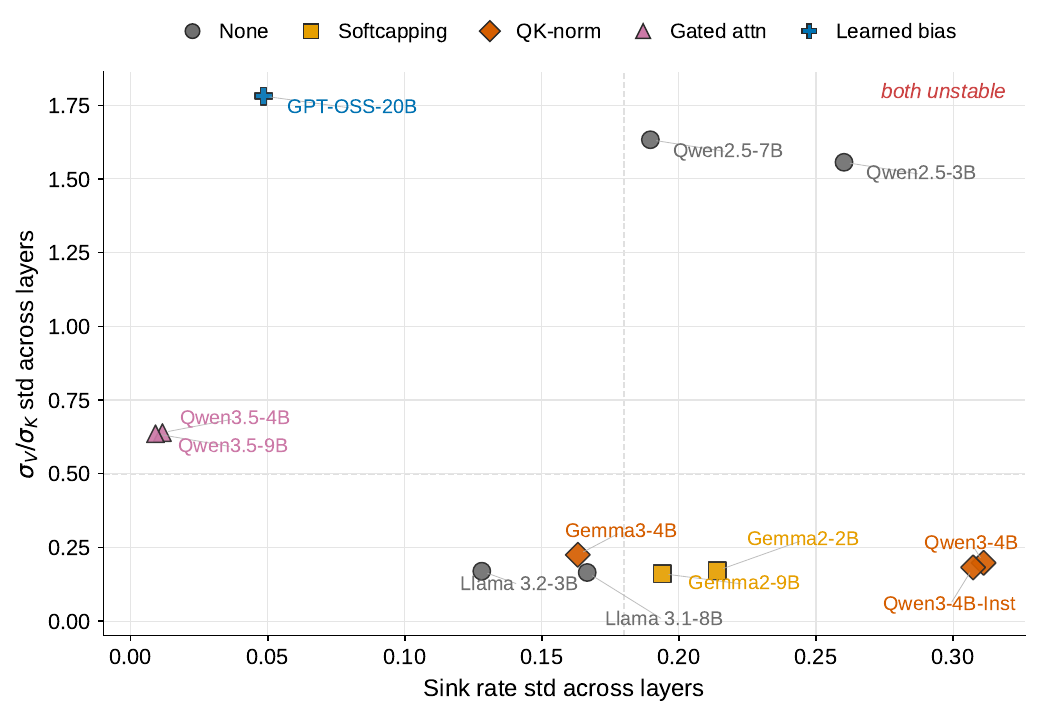}
\caption{Per-layer instability of sink rate (x-axis) vs.\
  $\sigma_V/\sigma_K$ (y-axis) across 11 models. Each point is one model.
  Qwen2.5 uniquely occupies the top-right quadrant where both signals
  are simultaneously unstable across layers.}
\label{fig:instability_scatter}
\end{figure}

Qwen2.5 stands out visually in both per-layer figures above. Its sink rate
oscillates sharply between high and low values across consecutive layers
(\cref{fig:sink_rate_per_layer}), a pattern unique to this family in our
evaluation. Its $\sigma_V/\sigma_K$ ratio crosses the key/value boundary
in both directions with extreme amplitude (\cref{fig:sigma_ratio_per_layer}):
Qwen2.5-3B ranges from 0.31 to 8.77 (std $= 1.56$) and Qwen2.5-7B from
0.30 to 6.81 (std $= 1.63$). \cref{fig:instability_scatter} summarizes
why this is unique: plotting per-layer sink rate std against $\sigma_V/\sigma_K$
std across all 11 models, Qwen2.5 is the \emph{only} family occupying the
high-high quadrant. Qwen3 has comparable sink rate variance (std $= 0.31$)
but stable $\sigma_V/\sigma_K$ (std $= 0.20$); GPT-OSS has high
$\sigma_V/\sigma_K$ variance (std $= 1.78$) but near-zero sink oscillation
(std $= 0.05$); crucially, its ratio remains consistently above 1 across
all layers, so the variance reflects magnitude differences rather than
regime instability, and \ourmethod{} reliably exploits the value-dominant
signal throughout. No other model has both instabilities simultaneously, and every eviction method relies on at least one consistent signal to identify important tokens; Qwen2.5 offers neither.


We also confirmed that Qwen 2.5 models have extreme grouped-query attention ratio. \cref{tab:gqa_ratios} shows the GQA ratios across all 11 models: Qwen2.5-3B and 7B have the highest ratios (8:1
and 7:1) among all-full-attention models. GPT-OSS-20B shares an 8:1 ratio but uses a 128-token sliding
window in half its layers, substantially reducing the diversity burden per KV
pair. For Qwen2.5, the high GQA compression means each cached KV pair must
serve many query heads simultaneously across the full context, leaving the
cache dense with non-redundant information and little room for token eviction
without significant information loss.

\begin{table}[h]
\centering\small
\caption{GQA compression ratios and effective attention scope across 11 models.
  Qwen2.5-3B and 7B have the highest GQA ratios among all-full-attention
  models; GPT-OSS shares an 8:1 ratio but limits the context burden via a
  128-token sliding window in half its layers.
}
\label{tab:gqa_ratios}
\begin{tabular}{llcccl}
\toprule
Model & Mechanism & Q heads & KV heads & GQA ratio & Attention scope \\
\midrule
Llama~3.1-8B  & None         & 32 & 8 & 4:1 & Full \\
Llama~3.2-3B  & None         & 24 & 8 & 3:1 & Full \\
Qwen2.5-3B    & None         & 16 & 2 & \textbf{8:1} & Full \\
Qwen2.5-7B    & None         & 28 & 4 & \textbf{7:1} & Full \\
\midrule
Gemma2-2B     & Softcapping  &  8 & 4 & 2:1 & Local 4096 (alt.) \\
Gemma2-9B     & Softcapping  & 16 & 8 & 2:1 & Local 4096 (alt.) \\
Qwen3-4B      & QK-norm      & 32 & 8 & 4:1 & Full \\
Gemma3-4B     & QK-norm      &  8 & 4 & 2:1 & Local 1024 (alt.) \\
\midrule
Qwen3.5-4B    & Gated attn   & 16 & 4 & 4:1 & Linear+full hybrid \\
Qwen3.5-9B    & Gated attn   & 16 & 4 & 4:1 & Linear+full hybrid \\
GPT-OSS-20B   & Learned sink & 64 & 8 & \textbf{8:1} & Local 128 (half layers) \\
\bottomrule
\end{tabular}
\end{table}

\section{Attention Ranking Stability Across Architectures}
\label{app:attn_rank_stability}

\subsection*{Motivation}

Attention-score eviction methods (TOVA~\cite{tova}, SnapKV~\cite{snapkv}, H$_2$O~\cite{h2o})
retain tokens with the highest observed attention weight.
Under block prompt processing, the eviction decision at block boundary $b$ is made
using only the attention computed from queries in blocks $1\ldots b$; queries in all later
blocks and all decode steps are unavailable. For this to work, the top-$k$ set identified
by queries in block $b$ must be a good proxy for what future queries will need.
We measure how stable the top-$k$ set actually is across different query positions
and whether this stability differs between sink-suppressed and non-suppressed models.

\subsection*{Measurement procedure}

For each model we run forward passes on 8 calibration sequences at context length 2048,
collecting the full attention weight matrix
$W^{(\ell)} \in \mathbb{R}^{H \times T \times T}$ at each layer $\ell$.
We then sample $n_q = 128$ query positions evenly spaced over the valid range
$t \in [k + 32,\, T-1]$ (requiring a KV prefix of at least $k + 32 = 288$ tokens
to form a meaningful top-$k$ set).

For each ordered pair of query positions $(t_i, t_j)$, we compare the top-$k$
attention sets over the \emph{shared} KV prefix of length $\min(t_i, t_j)$.
\begin{equation*}
  J(t_i, t_j, h, \ell)
  \;=\;
  \frac{
    \bigl|A^h_{t_i} \cap A^h_{t_j}\bigr|
  }{
    \bigl|A^h_{t_i} \cup A^h_{t_j}\bigr|
  },
  \qquad
  A^h_t = \mathrm{top}\text{-}k\bigl(W^{(\ell)}_{h,\,t,\,:c}\bigr),
  \quad c = \min(t_i, t_j),
\end{equation*}
where $k = 256$ (a representative eviction budget).
Using the shared prefix ensures both queries rank over the same token set.
For each query position $t_i$, we define its \emph{positional stability} as
\begin{equation*}
  s(t_i, h, \ell) \;=\; \frac{1}{n_q - 1} \sum_{j \neq i} J(t_i, t_j, h, \ell).
\end{equation*}
The model-level aggregate is the mean of $s(t_i, h, \ell)$ over all positions,
heads, and layers, averaged over the 8 calibration samples.

\subsection*{Results}

\cref{tab:attn_jaccard} reports the mean positional stability and its standard
deviation across layers for the 8 models evaluated.

\begin{table}[H]
\centering
\small
\caption{Mean top-256 attention Jaccard across query positions (model-level aggregate).
  Higher = more consistent rankings. Seven of nine models cluster in $[0.33, 0.40]$;
  Qwen3.5-4B falls slightly below (0.319) and GPT-OSS-20B is the sole high outlier (0.477).}
\label{tab:attn_jaccard}
\begin{tabular}{llcc}
\toprule
Model & Mechanism & Mean Jaccard & Std (across layers) \\
\midrule
Qwen2.5-7B   & None        & 0.334 & 0.023 \\
Qwen2.5-3B   & None        & 0.338 & 0.030 \\
Llama~3.2-3B & None        & 0.339 & 0.022 \\
Llama~3.1-8B & None        & 0.347 & 0.031 \\
\midrule
Qwen3-4B     & QK-norm     & 0.347 & 0.040 \\
Gemma2-9B    & Logit-cap   & 0.347 & 0.026 \\
Gemma3-4B    & QK-norm     & 0.398 & 0.036 \\
GPT-OSS-20B  & Learned sink & 0.477 & 0.141 \\
\midrule
Qwen3.5-4B   & Gated attn  & 0.319 & 0.030 \\
\bottomrule
\end{tabular}
\end{table}

\subsection*{Observed patterns}

\paragraph{Universal instability.}
Seven of nine models cluster tightly in $[0.33, 0.40]$ regardless of whether
attention is flattened. Qwen3.5-4B falls slightly below this range at 0.319,
consistent with near-zero attention sinks providing no stable high-mass anchor. The Jaccard ${\approx}0.4$ cited in the main body
(\cref{sec:observation}) is therefore not an artifact of a particular
architecture. It reflects a baseline property of attention itself.
Attention rankings over the full KV prefix are moderately inconsistent across
query positions in \emph{every} architecture tested.

\paragraph{Position-dependent decay.}
The aggregate Jaccard mixes positions with very different pool sizes.
At early query positions ($t \approx 288$--$714$), the shared KV prefix
is short and selecting 256 tokens out of $\sim\!300$ is nearly forced,
inflating Jaccard mechanically.
At later positions ($t \approx 1600$--$2034$), where the KV pool is large
and eviction is most consequential, the per-layer Jaccard falls to
$0.25$--$0.27$ on both non-suppressed (Llama~3.1-8B) and sink-suppressed (Qwen3-4B)
models.
This decay is consistent across architectures. The instability of attention
rankings grows with KV pool size, and is equally pronounced in the
sink-suppressed models.

\paragraph{GPT-OSS-20B outlier.}
GPT-OSS-20B registers a mean Jaccard of 0.477, substantially above the other models.
Its learned per-head sink logit absorbs the attention probability that would
otherwise concentrate on content-token sink positions, redistributing the
remaining probability across content tokens.
This redistribution makes the attention distribution closer to uniform over content
tokens, which has a dual effect. It removes any stable high-contrast anchor
(consistent with near-zero sink rate; \cref{tab:kv_diversity_full}), while also
making the top-$k$ set less sensitive to which query is used to evaluate it
(many tokens receive similar moderate weight, so the top-$k$ boundary is
relatively stable).
The high Jaccard therefore does not reflect \emph{reliable discriminative signal}
for eviction; it reflects that attention cannot distinguish which tokens matter,
a different form of failure.

\paragraph{No architecture-level separation.}
QK-norm, logit-capping, and no-suppression models all fall in the same Jaccard band.
This rules out the hypothesis that sink-suppressed models have \emph{worse}
ranking stability than sink-dominated ones. Instead, the data support a different
account. Attention rankings are universally unreliable at moderate to large KV pool
sizes. In sink-dominated models, TOVA is partially compensated by the sink. The
leading token(s) are in the top-$k$ of almost every query regardless of
content, providing a stable anchor. When sinks are suppressed, this anchor
disappears and attention-based eviction is left with only the unstable
content-token component, with no compensating mechanism.
This is consistent with the large performance gap on sink-suppressed
architectures (\cref{sec:ruler_results}). The gap reflects the \emph{loss of
compensation}, not an increase in ranking instability.

\section{Latency and Memory Sanity Check}
\label{app:latency_memory}

\begin{table}[t]
\centering\small
\caption{Long-context inference efficiency on H100 80\,GB (budget\,=\,2048).
  All eviction methods keep peak memory at $O(\text{budget})$ regardless of context length;
  \ourmethod{} variants match the cheapest eviction baseline (Sink) within 3--4\% on TTFT.}
\label{tab:latency_memory}
\resizebox{\linewidth}{!}{%
\begin{tabular}{llccccccc}
\toprule
 & & Dense & Sink & TOVA & SnapKV & KeyDiff & \ourmethod{} & V-Dir \\
\midrule
\multicolumn{9}{l}{\textit{Qwen3.5-4B (gated attention), ctx\,=\,128k}} \\
\quad Peak memory (GB) && 15.0 & 8.6 & 8.6 & 8.6 & 8.6 & 8.6 & 8.6 \\
\quad TTFT (s)         && 70.6 & \textbf{59.7} & 61.9 & 61.9 & 62.6 & 61.5 & 62.2 \\
\midrule
\multicolumn{9}{l}{\textit{GPT-OSS-20B (learned sink), ctx\,=\,65k}} \\
\quad Peak memory (GB) && 19.8 & 14.0 & 14.0 & 14.0 & 14.0 & 14.0 & 14.0 \\
\quad TTFT (s)         && 50.2 & \textbf{32.7} & 33.3 & 34.4 & 34.8 & 34.0 & 34.6 \\
\bottomrule
\end{tabular}}
\end{table}

The TTFT measurements characterize scoring overhead under the same block prompt
processing setup and cache budget across eviction methods. Under this setup,
eviction reduces peak memory relative to dense inference, and \ourmethod{}
stays within a few percent of the cheapest eviction baseline while
substantially improving retention on sink-suppressed models. Further
serving optimizations, including decode throughput and kernel-level cache
management, are complementary to the eviction scoring rule studied in this
paper.

\section{MATH-500 Dense Decode Lengths}
\label{app:math500_dense}

%

\begin{table}[t]
\centering
\caption{MATH-500 dense baselines ($n\!=\!100$, first 100 problems). For all models except
  Qwen3.5, $k\!=\!5$ generations per problem, flex@$k$ = mean \texttt{grade\_answer} accuracy, and
  score = pass@$k$ ($\geq$1 correct). For Qwen3.5$^\star$, greedy $k\!=\!1$ with thinking
  disabled; flex@$k$ and score both equal pass@1. Tokens and empty-response rate are
  per-generation. \emph{Budget} is \texttt{max\_new\_tokens}; Trunc\% is the fraction of
  generations that hit it. $^\ddagger$deepseek-r1-distill-llama-8b aggregated from
  \texttt{dense\_decoded/} after a BPE-marker post-processing fix.
  $^\star$Qwen3.5 uses \texttt{enable\_thinking=False}, greedy (\texttt{do\_sample=False}).
  Average token counts for Qwen3.5 set the reference lengths for eviction budgets
  $b\!=\!256$ and $b\!=\!512$ in \cref{tab:math500_eviction}.}
\label{tab:math500_dense}
\resizebox{\linewidth}{!}{%
\begin{tabular}{llcrrrrrr}
\toprule
Category & Model & Budget & flex@$k$ & exact@$k$ & score & Avg tok & Empty\% & Trunc\% \\
\midrule
\multirow{2}{*}{Logit softcapping}   & Gemma2-2B                         & 16k & 0.216 & 0.176 & 0.430 & 572    & 45.2 & 0.0 \\
                                     & Gemma2-9B                         & 16k & 0.526 & 0.446 & 0.680 & 430    & 16.6 & 0.0 \\
\midrule
\multirow{3}{*}{QK-normalization}    & Gemma3-4B                         & 16k & 0.742 & 0.682 & 0.820 & 932    & 11.2 & 0.0 \\
                                     & Qwen3-4B            & 16k & 0.928 & 0.792 & 0.960 & 1\,947 & 1.4  & 0.0 \\
                                     & Qwen3-4B-Thinking-2507            & 32k & 0.888 & 0.704 & 0.940 & 6\,974 & 6.6  & 0.0 \\
\midrule
\multirow{2}{*}{Gated hybrid}        & Qwen3.5-4B$^\star$                & 32k & 0.890 & 0.780 & 0.890 & 1\,124 & 0.0  & 0.0 \\
                                     & Qwen3.5-9B$^\star$                & 32k & 0.890 & 0.790 & 0.890 & 1\,228 & 0.0  & 0.0 \\
\midrule
Learned attention sink               & GPT-OSS-20B (reasoning=high)      & 16k & 0.934 & 0.772 & 0.960 & 2\,432 & 1.4  & 5.0 \\
\midrule
\multirow{2}{*}{Reasoning distill}   & DS-R1-Distill-Qwen-1.5B           & 32k & 0.836 & 0.636 & 0.930 & 3\,708 & 8.0  & 0.0 \\
                                     & DS-R1-Distill-Llama-8B$^\ddagger$ & 32k & 0.834 & 0.624 & 0.920 & 4\,147 & 4.2  & 0.0 \\
\bottomrule
\end{tabular}}
\end{table}

\paragraph{Standard sampling protocol.}
For all models except Qwen3.5, we sample $k{=}5$ generations per problem using the temperature and top-$p$ from each model's official card, with a maximum of 16k or 32k new tokens as shown in \cref{tab:math500_dense}. Cache eviction budgets are set to approximately 25\% and 50\% of the average dense generation length, rounded to clean multiples of 256/512 tokens for Gemma3-4B and Qwen3.5, 512/1024 tokens for Qwen3-4B and GPT-OSS-20B, and 1024/2048 tokens for the R1-Distill models.

\paragraph{Qwen3.5 no-think greedy protocol.}
With \texttt{enable\_thinking=True} ($k{=}5$, recommended sampling parameters),
the dense pilot on 100 problems yielded average generation
lengths of 21,620 tokens (4B) and 18,611 tokens (9B) with a 32k
\texttt{max\_new\_tokens} cap, and 39\% and 35\% of generations were truncated
at that limit. For context, every other model in this evaluation averages
932--4,147 tokens per generation (see \cref{tab:math500_dense}), making
Qwen3.5 think-on traces 5--18$\times$ longer than the next longest model.
Even at a 6-hour per-sample generation timeout, many traces
did not complete. 

Beyond the engineering obstacle, the truncation severely degrades accuracy:
a generation cut off mid-reasoning cannot produce a correctly-formatted
\texttt{\textbackslash boxed\{\}} answer, inflating the empty-response rate to
19\% (4B) and 15.4\% (9B). The resulting think-on dense baselines of
flex@5\,=\,0.790 (4B) and 0.828 (9B) are suppressed by this artifact and
understate the model's actual capability. We confirmed that
\texttt{enable\_thinking=False} outperforms \texttt{enable\_thinking=True}
on MATH-500 for both model sizes.

We therefore evaluated Qwen3.5 with thinking disabled
(\texttt{enable\_thinking=False}), greedy decoding (\texttt{do\_sample=False},
temperature 0), and $k{=}1$. Despite the setting name, \emph{no-think mode
still generates step-by-step mathematical reasoning}: the model works through
problems with structured intermediate steps in its response. 
The resulting responses are complete solutions averaging 1,124 tokens
(4B) and 1,228 tokens (9B) with zero truncation. The no-think dense baseline
of flex@1\,=\,0.890 for both models is 10.0\,pp (4B) and 6.2\,pp (9B) higher
than the truncation-affected think-on baseline, making it the more reliable
reference for measuring eviction retention.

The eviction budgets $b{=}256$ and $b{=}512$ correspond to approximately
23\% and 46\% of the no-think dense length for the 4B model, and 21\% and
42\% for the 9B model.
Retention in \cref{tab:math500_eviction} is computed relative to the no-think greedy dense baseline for Qwen3.5, and relative to the $k{=}5$ flex@5 baseline for all other models.

\section{Sink Formatting Collapse on Qwen3.5-4B}
\label{app:sink_formatting}

Sink eviction on Qwen3.5-4B produces near-zero flex\_match at both budgets
(1.1\% at $b{=}256$, 31.5\% at $b{=}512$), far below every other method.
We analyze all 100 MATH-500 generations at each budget to identify the failure mechanism.

\paragraph{Math is largely correct; output format is not.}
Of the 99 failures at $b{=}256$, 67\% contain the ground-truth answer string somewhere in the response (stripping whitespace and \LaTeX\ delimiters); at $b{=}512$ this rises to 74\%.
The model solves the problem but fails to wrap the final answer in \texttt{\textbackslash boxed\{\}}, which is required for flex\_match scoring.
\cref{tab:sink_failure_cats} breaks down the failure categories.

\begin{table}[t]
\centering\small
\caption{Failure categories for Sink eviction on Qwen3.5-4B, MATH-500 ($n{=}100$).
``Format (no box)'' = correct reasoning, answer present but not in \texttt{\textbackslash boxed\{\}}.
``Truncated'' = generation ends mid-reasoning without conclusion.
``Context lost'' = model states the problem was not provided (input fidelity failure).
``Wrong answer'' = \texttt{\textbackslash boxed\{\}} present but mathematically incorrect.}
\label{tab:sink_failure_cats}
\begin{tabular}{lrrrr}
\toprule
Category & b256 & \% & b512 & \% \\
\midrule
Format (no \texttt{\textbackslash boxed\{\}}) & 59 & 60\% & 40 & 56\% \\
Truncated mid-reasoning           & 37 & 37\% & -- & -- \\
Context lost (input fidelity)     & 25 & 25\% & $<$5 & -- \\
Wrong answer (\texttt{\textbackslash boxed\{\}} present) & 2 & 2\% & 1 & 1\% \\
\bottomrule
\end{tabular}
\end{table}

\paragraph{Three failure categories (not mutually exclusive).}
\textbf{(1) Format regression} (60\% of b256 failures). The model produces a fully correct solution ending with \texttt{**Final Answer:** \$\ldots\$} or \texttt{Answer: \$\ldots\$} but omits \texttt{\textbackslash boxed\{\}}, which is required for flex\_match scoring.

\textbf{(2) Truncated generation} (37\% of b256 failures). The response ends abruptly mid-reasoning, without a concluding answer line. The model's state for tracking generation progress is degraded by aggressive context compression, producing premature termination.

\textbf{(3) Context collapse} (25\% of b256 failures). The model outputs ``\textit{It appears that the math problem you intended to ask about was not included in your message. Please provide the problem statement.}'' The KV cache at $b{=}256$ (25\% retention) is too compressed to reconstruct the question, which is a pure input-fidelity failure.

\paragraph{Contrast with other b256 methods.}
FastCAOTE, TOVA, and SnapKV also show low flex\_match at b256 (34.8\%, 39.3\%, 24.7\% respectively), and their failures also exhibit format regression and truncation at similar rates (55--68\% of failures contain the correct answer).
However, all three \emph{recover strongly at b512} (83.1\%, 80.9\%, 84.3\%), demonstrating that their b256 failures are budget-pressure effects. A 25\% retention budget is simply too little capacity for reliable formatting and extended reasoning chains.
Sink's failure is qualitatively different. It persists at b512 (31.5\%), indicating that the problem is not resolved by increasing budget alone.


\end{document}